\documentclass[10pt]{article} % For LaTeX2e
\pdfoutput=1
\usepackage[accepted]{tmlr}
\usepackage{amsmath,amsfonts,bm}

\def\eqref#1{equation~\ref{#1}}
\def\1{\bm{1}}

\DeclareMathAlphabet{\mathsfit}{\encodingdefault}{\sfdefault}{m}{sl}
\SetMathAlphabet{\mathsfit}{bold}{\encodingdefault}{\sfdefault}{bx}{n}

\usepackage{hyperref}
\usepackage{array}
\usepackage[caption=false,textfont=sf]{subfig}
\usepackage{textcomp}
\usepackage{enumitem}
\usepackage{xcolor}
\usepackage{booktabs}
\usepackage{nicefrac}
\usepackage{microtype}
\usepackage{scalerel}
\usepackage{multirow}
\usepackage{lipsum}
\usepackage{wrapfig}
\usepackage{bbm}
\usepackage{stfloats}
\usepackage{verbatim}
\usepackage{graphicx}
\usepackage{url}

\usepackage{amsmath,amsfonts}
\usepackage{amssymb}
\usepackage{mathtools}
\usepackage{amsthm}
\usepackage{breqn}

\usepackage{algorithm}
\usepackage{algpseudocode}

\theoremstyle{plain}
\newtheorem{theorem}{Theorem}%[section]

\newtheorem{lemma}[theorem]{Lemma}

\theoremstyle{definition}

\theoremstyle{remark}

\newcommand\independent{\protect\mathpalette{\protect\independenT}{\perp}}
\def\independenT#1#2{\mathrel{\rlap{$#1#2$}\mkern2mu{#1#2}}}

\allowdisplaybreaks

\title{Incorporating Prior Knowledge into Neural Networks \\ through an Implicit Composite Kernel}

\author{\name Ziyang Jiang \email ziyang.jiang@duke.edu \\
\addr Department of Civil and Environmental Engineering \\
Duke University
\AND
\name Tongshu Zheng \email tongshuzheng92@gmail.com \\
\addr Division of Natural and Applied Science \\
Duke Kunshan University
\AND
\name Yiling Liu \email yiling.liu@duke.edu \\
\addr Program in Computational Biology and Bioinformatics \\
Duke University School of Medicine
\AND
\name David Carlson \email david.carlson@duke.edu \\
\addr Department of Civil and Environmental Engineering \\
Department of Biostatistics and Bioinformatics\\
Department of Computer Science\\
Duke University
}

\def\month{02}  % Insert correct month for camera-ready version
\def\year{2024} % Insert correct year for camera-ready version
\def\openreview{\url{https://openreview.net/forum?id=HhjSalvWVe}} % Insert correct link to OpenReview for camera-ready version

\begin{document}

\maketitle

\begin{abstract}
It is challenging to guide neural network (NN) learning with prior knowledge. In contrast, many known properties, such as spatial smoothness or seasonality, are straightforward to model by choosing an appropriate kernel in a Gaussian process (GP). Many deep learning applications could be enhanced by modeling such known properties. For example, convolutional neural networks (CNNs) are frequently used in remote sensing, which is subject to strong seasonal effects. We propose to blend the strengths of NNs and the clear modeling capabilities of GPs by using a composite kernel that combines a kernel implicitly defined by a neural network with a second kernel function chosen to model known properties (e.g., seasonality). \textcolor{black}{We implement this idea by combining a deep network and an efficient mapping function based on either Nystr\"{o}m approximation or random Fourier features, which we call Implicit Composite Kernel (ICK).} We then adopt a sample-then-optimize approach to approximate the full GP posterior distribution. We demonstrate that ICK has superior performance and flexibility on both synthetic and real-world datasets including a remote sensing dataset. The ICK framework can be used to include prior information into neural networks in many applications.
\end{abstract}

\section{Introduction}
\label{sec:1}

In complex regression tasks, input data often contains \emph{multiple sources of information}. These sources can be presented in both high-dimensional (e.g. images, audios, texts, etc.) and low-dimensional (e.g. timestamps, spatial locations, etc.) forms. A common approach to learn from high-dimensional information is to use neural networks (NNs) \citep{goodfellow2016deep, lecun2015deep}, as NNs are powerful enough to capture the relationship between complex high-dimensional data and target variables of interest. In many areas, NNs are standard practice, such as the dominance of Convolutional Neural Networks (CNNs) for image analysis \citep{jiang2022improving, zheng2021local, zheng2020estimating}. In contrast, for low-dimensional information, we usually have some prior knowledge on how the information relates to the predictions. As a concrete example, consider a remote sensing problem where we predict ground measurements from satellite imagery with associated timestamps. \textit{A priori}, we expect the ground measurements to vary periodically with respect to time between summer and winter due to seasonal effects.  We would typically use a CNN to capture the complex relationship between the imagery and the ground measurements. In this case, we want to guide the learning of the CNN with our prior knowledge about the seasonality. This is challenging because knowledge represented in NNs pertains mainly to correlation between network units instead of quantifiable statements \citep{marcus2018deep}.

Conversely, Gaussian processes (GPs) have been used historically to incorporate relevant prior beliefs by specifying the appropriate form of its kernel (or covariance) function \citep{bishop2006pattern, williams2006gaussian}. One approach to modeling multiple sources of information is to assign a relevant kernel function to each source of information respectively and combine them through addition or multiplication, resulting in a \textit{composite kernel function} \citep{duvenaud2014automatic}. This formulation means that specifying a kernel to match prior beliefs on one source of information is straightforward. Such composite kernel learning techniques are extensively used in many application areas such as multi-media data \citep{mcfee2011learning}, neuroimaging \citep{zhang2011multimodal}, spatial data analysis, and environmental data analysis \citep{kim2005analyzing, petelin2013evolving}. In view of the clear modeling capabilities of GP, it is desirable to examine how a NN could be imbued with the same modeling ease.

In recent years, researchers have come up with a variety of methods to incorporate prior knowledge into NNs. These efforts can be broken into many categories, such as those that add prior information through loss terms like physics-informed NNs \citep{lagaris1998artificial, moseley2020solving}.  Here, we focus on the major category of those methods that build integrated models of NNs and GPs with various structures \citep{van2017convolutional, wilson2016deep, wilson2011gaussian}. Related to our proposed methodology, \citet{pearce2020expressive} exploited the fact that a Bayesian neural network (BNN) approximates a GP to construct additive and multiplicative kernels, but they were limited to specific predefined kernels. \citet{matsubara2020ridgelet} then resolved this limitation by constructing priors of BNN parameters based on the ridgelet transform and its dual, but they did not explicitly show how their approach works for data with multiple sources of information. To our knowledge, no existing approach allows a modeler to choose any appropriate kernel over multiple sources.

We address this limitation by presenting a simple yet novel Implicit Composite Kernel (ICK) framework, which processes high-dimensional information using a kernel implicitly defined by a neural network and low-dimensional information using a chosen kernel function. The low-dimensional kernels are mapped into the neural network framework to create a straightforward and simple-to-learn implementation. Our key results and contributions are:
\begin{itemize} [leftmargin=*]
\item We analytically show our ICK framework, under reasonable assumptions, is approximately equivalent to sampling from a Gaussian process regression (GPR) model with a composite kernel \emph{a priori}.
\item We adopt a sample-then-optimize procedure to ICK to approximate the full posterior distribution of a GP with a composite kernel.
\item We show that ICK yields better performance on prediction and forecasting tasks, even with limited data.
\item We show that ICK can flexibly capture the patterns of the low-dimensional information without bespoke pre-processing procedures or complex NN structures.
\end{itemize}

Based on these contributions, we believe ICK is useful in learning from complex \emph{hybrid} data with prior knowledge, especially in remote sensing and spatial statistics.

\section{Related Work}
\label{sec:2}

\paragraph{Equivalence between NNs and GPs} 
The equivalence between GPs and randomly initialized single-layer NNs with infinite width was first shown by \citet{neal1996priors}. With the development of modern deep learning, researchers further extended this relationship to deep networks \citep{lee2017deep, matthews2018gaussian} and convolutional neural networks (CNNs) \citep{garriga2018deep,novak2018bayesian}. This relationship is crucial for showing the resemblance between GPR and our ICK framework, which is discussed in Section \ref{sec:4.1}.

\paragraph{NNs with prior knowledge}
As mentioned before, one approach to equip NNs with prior knowledge is to modify the loss function. For example, \citet{lagaris1998artificial} solved differential equations (DEs) using NNs by setting the loss to be a function whose derivative satisfies the DE conditions. Another approach is to build integrated models of NNs and kernel-based models. For example, \citet{wilson2011gaussian} implemented a regression network with GP priors over latent variables and made inference by approximating the posterior using Variational Bayes or sampling from the posterior using Gibbs sampling scheme. \citet{garnelo2018neural} introduced a class of neural latent variable models called Neural Processes (NPs) which are capable of learning efficiently from the data and adapting rapidly to new observations. \textcolor{black}{In addition to these, various studies have explored the integration of Neural Networks (NNs) and kernel methods \citep{hinton2007using,wilson2016stochastic,adlam2020exploring}.} Our ICK framework fuses prior knowledge into NNs by modulating the learnt features using another set of features outputted from a kernel-based mapping, which can also be viewed as an integrated model of NNs and kernel machines.

\paragraph{GP with composite kernels} 
Composite kernel GPs are widely used in both machine learning \citep{duvenaud2014automatic, williams2006gaussian} and geostatistical modeling \citep{datta2016hierarchical, gelfand2016spatial}. GPR in geostatistical modeling is also known as \emph{kriging} \citep{journel1976mining, krige1951statistical}, which serves as a surrogate model to replace expensive function evaluations. The inputs for a composite GP are usually low-dimensional (e.g. spatial distance) as GPs do not scale well with the number of samples for high-dimensional inputs \citep{bouhlel2019gradient, bouhlel2016improving}. To overcome this issue, \citet{pearce2020expressive} and \citet{matsubara2020ridgelet} developed BNN analogues for composite GPs. Similar to these studies, our ICK framework can also be viewed as a simulation for composite GPs. 

\paragraph{GP for large datasets} 
\textcolor{black}{Since training and inference of exact GP scales $\mathcal{O}(N^3)$, either parallel computing \cite{wang2019exact,adlam2023kernel} or kernel approximation are needed to scale GP to large datasets.} Nystr\"{o}m low-rank matrix approximation \citep{drineas2005nystrom, williams2000using} and Random Fourier Features \citep{rahimi2007random, rahimi2008weighted} are two commonly used approximation methods. \textcolor{black}{Building upon these concepts, several popular frameworks, such as sparse GPs \citep{snelson2005sparse, titsias2009variational, hensman2013gaussian}, have been developed to facilitate GP inference on large datasets.} In our research, we draw inspiration from these approximation methods and utilize them as \emph{transformation functions} to project the kernel matrix into latent space representations, as discussed in Section \ref{sec:4.2}.

\section{Background}
\label{sec:3}

\subsection{Problem Setup}
\label{sec:3.1}
To formalize the problem, we have a training data set which contains $N$ data points $\boldsymbol{X} = [\boldsymbol{x}_i]_{i=1}^{N} = [\boldsymbol{x}_1, \boldsymbol{x}_2, ..., \boldsymbol{x}_N]^{T}$ and the corresponding labels of these data points are $\boldsymbol{y} = [y_i]_{i=1}^{N} = [y_1, y_2, ..., y_N]^{T}$ where $y_i \in \mathbb{R}$. Each data point $\boldsymbol{x}_i = \{\boldsymbol{x}_i^{(1)}, \boldsymbol{x}_i^{(2)}, ..., \boldsymbol{x}_i^{(M)}\}$ is composed of information from $M$ different sources where the $m^{th}$ source of information of the $i^{th}$ data point is denoted as $\boldsymbol{x}_i^{(m)} \in \mathbb{R}^{D_m}$. Our goal is to learn a function $\hat{y}_i = f(\boldsymbol{x}_i): \mathbb{R}^{D_1} \times \mathbb{R}^{D_2} \times ... \times \mathbb{R}^{D_M} \rightarrow \mathbb{R}$ which takes in a data point $\boldsymbol{x}_i$ and outputs a predicted value $\hat{y}_i$.

\subsection{Composite GPs}
\label{sec:3.2}
A Gaussian process (GP) describes a distribution over functions \citep{williams2006gaussian}. A key property of GP is that it is completely defined by a mean function $\mu(\boldsymbol{x})$ and a kernel function $K(\boldsymbol{x}, \boldsymbol{x}')$ \textcolor{black}{where $\boldsymbol{x}$ and $\boldsymbol{x}'$ represent different samples from the training dataset}. The mean function $\mu(\boldsymbol{x})$ is often assumed to be zero for simplicity. In that case, the outcome function is
\begin{equation}
f(\boldsymbol{x}) \sim \mathcal{GP}\left(0, K(\boldsymbol{x}, \boldsymbol{x}')\right).
\end{equation}
Any finite subset of random variables has a multivariate Gaussian distribution with mean $\boldsymbol{0}$ and kernel matrix $\boldsymbol{K}$ whose entries can be calculated as $\boldsymbol{K}_{ij} = K(\boldsymbol{x}_i, \boldsymbol{x}_j)$ where $1 \leq i, j \leq N$. In many situations, the full kernel function is built by a composite kernel by combining simple kernels through addition $K^{\text{comp}}(\boldsymbol{x}, \boldsymbol{x}') = K_1(\boldsymbol{x}, \boldsymbol{x}') + K_2(\boldsymbol{x}, \boldsymbol{x}')$ or multiplication $K^{\text{comp}}(\boldsymbol{x}, \boldsymbol{x}') = K_1(\boldsymbol{x}, \boldsymbol{x}') K_2(\boldsymbol{x}, \boldsymbol{x}')$ \citep{duvenaud2014automatic}. A useful property that ICK exploits is that $K_1$ and $K_2$ can take different subparts of $\boldsymbol{x}$ as their inputs. For example, $K^{\text{comp}}(\boldsymbol{x}, \boldsymbol{x}') = K_1 (\boldsymbol{x}^{(1)},{\boldsymbol{x}^{(1)}}') + K_2 (\boldsymbol{x}^{(2)},{\boldsymbol{x}^{(2)}}')$ or $K^{\text{comp}}(\boldsymbol{x}, \boldsymbol{x}') = K_1 (\boldsymbol{x}^{(1)},{\boldsymbol{x}^{(1)}}') K_2 (\boldsymbol{x}^{(2)},{\boldsymbol{x}^{(2)}}')$. 

\subsection{Correspondence between GPs and NNs}
\label{sec:3.3}
\citet{neal1996priors} proved that a single-hidden layer network with infinite width is \emph{exactly equivalent} to a GP over data indices $i = 1, 2, ..., N$ under the assumption that the weight and bias parameters of the hidden layer are i.i.d. Gaussian with zero mean. This statement was then extended to deep NNs \citep{lee2017deep, matthews2018gaussian} and convolutional NNs \citep{garriga2018deep,novak2018bayesian}. Specifically, let $\boldsymbol{z} = f_{\text{NN}}\left(\boldsymbol{x}^{(1)}\right): \mathbb{R}^{D_1} \rightarrow \mathbb{R}^{p}$ be the latent representation extracted from $\boldsymbol{x}^{(1)}$ where $p$ is the dimension of the extracted representation and $f_{\text{NN}}$ is a neural network with zero-mean i.i.d. parameters \textcolor{black}{and continuous activation function $\phi$ which satisfies the linear envelope property}
\begin{equation}
|\phi(u)| \leq c + m|u| \quad \forall u \in \mathbb{R},
\label{eq:2}
\end{equation}
\textcolor{black}{if there exists $c, m \geq 0$. 
 This includes many standard nonlinearities.} The $k^{th}$ entry of this representation will \textcolor{black}{converge in distribution to} a NN-implied GP \textcolor{black}{in the infinite width limit}
\begin{equation}
f_{\text{NN}}(\boldsymbol{x}^{(1)})_k \xrightarrow[]{d} \mathcal{GP}(0, K^{\text{NN}}( \boldsymbol{x}^{(1)}, {\boldsymbol{x}^{(1)}}' )).
\label{eq:3}
\end{equation}
\textcolor{black}{Here $K^{\text{NN}}$ in Equation \ref{eq:3} denotes the covariance function of the equivalent GP as the width of $f_{\text{NN}}$ goes to infinity and can be computed numerically in a recursive manner \citep{lee2017deep}.} That is to say, the $k^{th}$ component $z_k$ of the representation extracted by the network has zero mean $\mathbb{E}_{p(\boldsymbol{\theta}^{(1)})}\left[z_{ik}^{(1)}\right] = 0$ for all $i = 1, 2, ..., N$ where $\theta$ represents the network parameters. The covariance between $z_{ik}^{(1)}$ and $z_{jk}^{(1)}$ for \emph{different} data indices $i, j = 1, 2, ..., N$ can be approximated as $\text{cov} ( z_{ik}^{(1)},z_{jk}^{(1)} ) = \mathbb{E}_{p(\boldsymbol{\theta}^{(1)})}\left[ z_{ik}^{(1)}z_{jk}^{(1)} \right] \approx K^{\text{NN}} \left( \boldsymbol{x}_i^{(1)}, \boldsymbol{x}_j^{(1)} \right)$ where $\boldsymbol{x}_i^{(1)}$ and $\boldsymbol{x}_j^{(1)}$ are the corresponding inputs \textcolor{black}{in case the network width is finite.}

\begin{figure}[t!]
\begin{minipage}[c]{0.61\textwidth}
\includegraphics[width=\textwidth]{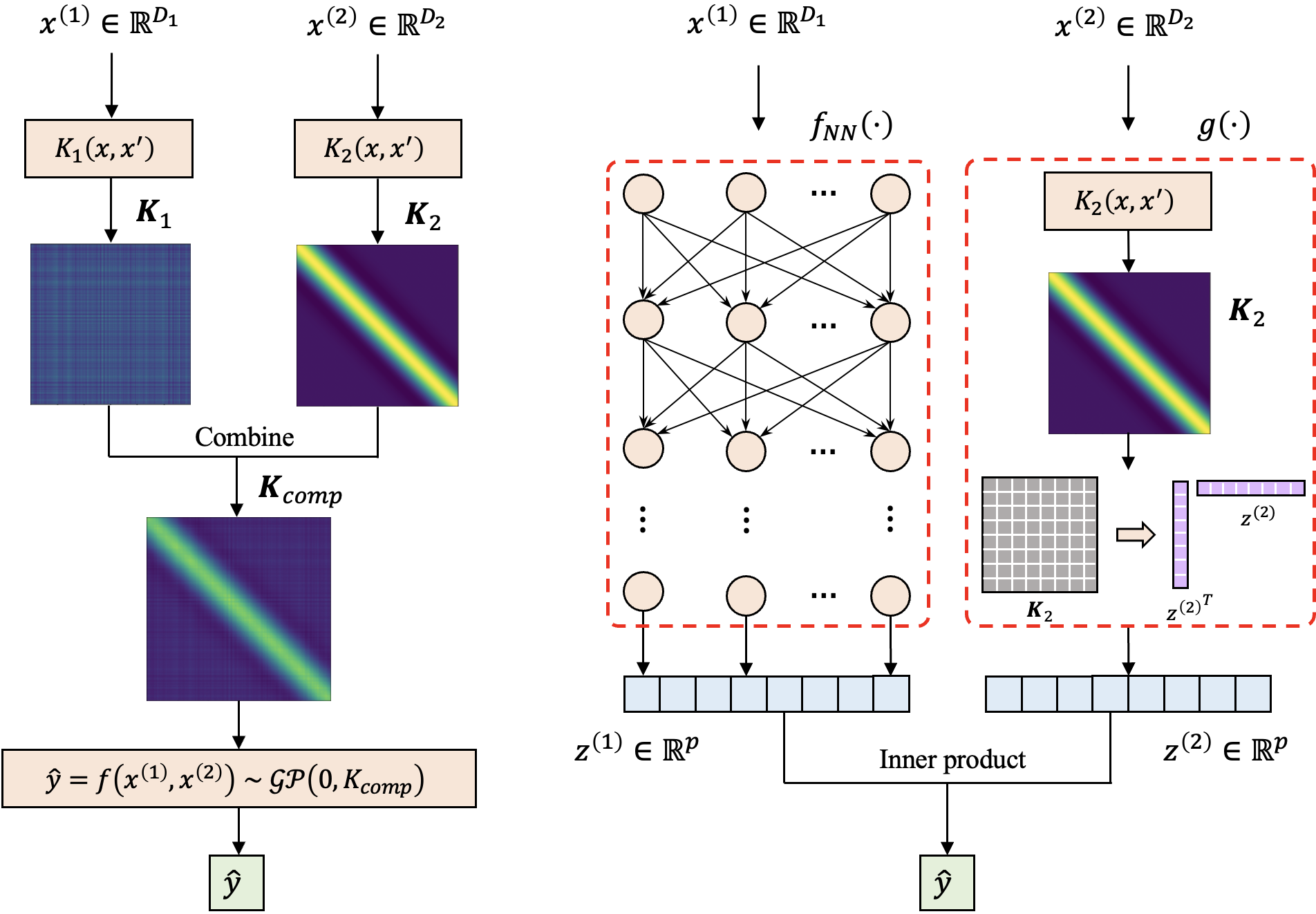}
\end{minipage}\hfill
\begin{minipage}[c]{0.35\textwidth}
\caption{Given data containing 2 sources of information $\boldsymbol{x}^{(1)}$ and $\boldsymbol{x}^{(2)}$, we can process the data using either (\textbf{Left}) a composite Gaussian process regression (GPR) model or (\textbf{Right}) our ICK framework where $\boldsymbol{x}^{(1)}$ is processed with a neural network $f_{\text{NN}}(\cdot)$ and $\boldsymbol{x}^{(2)}$ is processed with $g(\cdot)$ where $g(\cdot)$ consists of a kernel function $K_2$ and some transformation which maps the kernel matrix $\boldsymbol{K}_2$ into the latent space.} 
\label{fig:1}
\end{minipage}
\end{figure}

\section{Implicit Composite Kernel}
\label{sec:4}

We show the structure of a composite GPR model and our ICK framework in Figure \ref{fig:1}. To make the illustration clear, we limit ourselves to data with information from 2 different sources $\boldsymbol{x} = \left\{ \boldsymbol{x}^{(1)}, \boldsymbol{x}^{(2)} \right\}$ where $\boldsymbol{x}^{(1)}$ is high-dimensional and $\boldsymbol{x}^{(2)}$ is low-dimensional (i.e. $D_1 \gg D_2$) with some known relationship with the target $y$. We are inspired by composite GPR, which computes 2 different kernel matrices $\boldsymbol{K}_1$ and $\boldsymbol{K}_2$ and then combines them into a single composite kernel matrix $\boldsymbol{K}^{\text{comp}}$. However, as discussed before, it is more suitable to use a NN to learn from the high dimensional information $\boldsymbol{x}^{(1)}$. In our ICK framework, we process $\boldsymbol{x}^{(1)}$ with a NN $f_{\text{NN}}(\cdot): \mathbb{R}^{D_1} \rightarrow \mathbb{R}^p$ and $\boldsymbol{x}^{(2)}$ with a mapping $g(\cdot): \mathbb{R}^{D_2} \rightarrow \mathbb{R}^p$, which consists of a kernel function $K_2$ followed by a kernel-to-latent-space transformation (described in Section \ref{sec:4.2}), resulting in two latent representations $\boldsymbol{z}^{(1)}, \boldsymbol{z}^{(2)} \in \mathbb{R}^p$. Then, we make a prediction $\hat{y}$ by doing an \emph{inner product} between these two representations $\hat{y} = f_{\text{NN}}\left(\boldsymbol{x}^{(1)}\right) \cdot g\left(\boldsymbol{x}^{(2)}\right)$. Finally, the parameters of both the NN and the kernel function are learned via gradient-based optimization methods \citep{bottou2018optimization}.

\begin{figure}[t!]
\begin{minipage}[c]{0.61\textwidth}
\includegraphics[width=\textwidth]{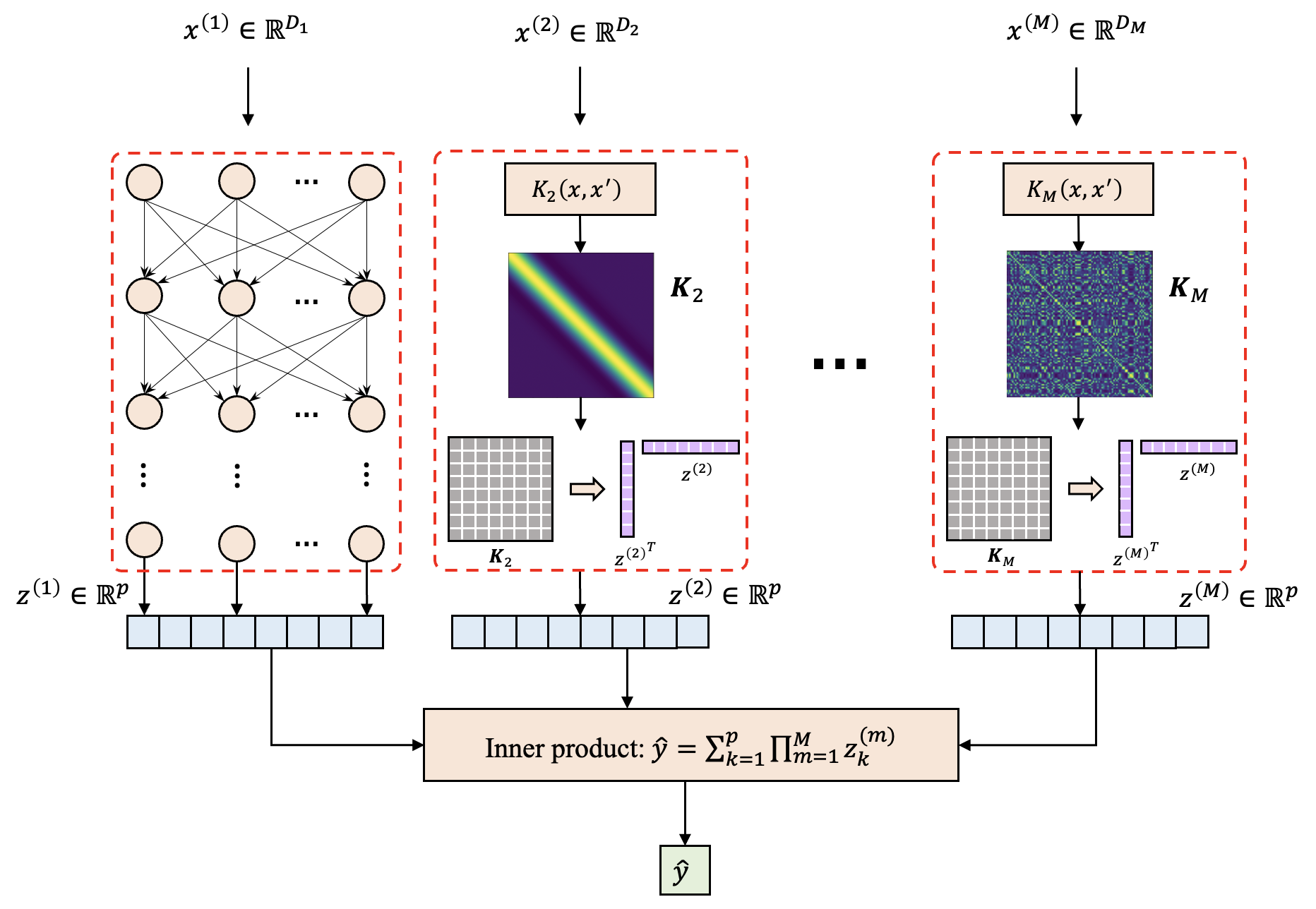}
\end{minipage}\hfill
\begin{minipage}[c]{0.35\textwidth}
\caption{Given data containing $M$ sources of information $\boldsymbol{x} = \left\{ x^{(1)}, x^{(2)}, ..., x^{(M)} \right\}$, we can process the data using our ICK framework where high-dimensional information (e.g. $x^{(1)}$ in the figure) is processed using a neural network and low-dimensional information (e.g. $x^{(2)}$ in the figure) is processed using a kernel function and some transformation which maps the kernel matrix into the latent space.} 
\label{fig:2}
\end{minipage}
\end{figure}

\textcolor{black}{Besides the formulation in Figure \ref{fig:1}, ICK can also process data $\boldsymbol{x} = \left\{ x^{(1)}, x^{(2)}, ..., x^{(M)} \right\}$ with $M > 2$ sources of information as shown in Figure \ref{fig:2}. Here $K_1, ..., K_M$ represent different types of kernels with trainable parameters. The final prediction is calculated by a chained inner product of all extracted representations $\hat{y} = \sum_{k=1}^{p} \prod_{m=1}^{M} z_k^{(m)}$.}

In the sections below, we first show the relationship between ICK and a composite GPR model with a multiplicative kernel \emph{a priori} in Section \ref{sec:4.1}, which is used to \textbf{motivate the model form}. We then show how we implement the kernel-to-latent-space transformation in Section \ref{sec:4.2}. \textcolor{black}{To enable a full GP posterior \emph{approximation} for ICK, we will provide insight into how we derive uncertainty estimates from ICK. This can be achieved either through direct variance calculation (Section \ref{sec:4.3.1}) or through the utilization of an ensemble approach (Section \ref{sec:4.3.2}).}

\subsection{Resemblance between Composite GPR and ICK}
\label{sec:4.1}
We will analytically prove the following theorem for data with information from 2 different sources $\boldsymbol{x} = \left\{ \boldsymbol{x}^{(1)}, \boldsymbol{x}^{(2)} \right\}$ for clarity, and we note this theorem can be straightforwardly extended to $M > 2$.
\begin{theorem}
\label{thm:1}
Let $f_{\text{NN}}: \mathbb{R}^{D_1} \rightarrow \mathbb{R}^p$ be a NN function with random weights and $g: \mathbb{R}^{D_2} \rightarrow \mathbb{R}^p$ be a mapping function, and define an inner product between the representations $\hat{y} = f_{\text{ICK}} \left( \boldsymbol{x}^{(1)}, \boldsymbol{x}^{(2)} \right) = f_{\text{NN}}\left(\boldsymbol{x}^{(1)}\right) \cdot g\left(\boldsymbol{x}^{(2)}\right) = {\boldsymbol{z}^{(1)}}^T \boldsymbol{z}^{(2)}$. \textcolor{black}{Then $f_{\text{ICK}}$ will converge in distribution to a GP in the infinite width limit a priori}
\begin{equation}
f_{\text{ICK}} \xrightarrow[]{d} \mathcal{GP} ( 0, K^{\text{NN}}_1 (\boldsymbol{x}^{(1)}, {\boldsymbol{x}^{(1)}}') K_2 (\boldsymbol{x}^{(2)}, {\boldsymbol{x}^{(2)}}') ), 
\end{equation}
if \textcolor{black}{$f_{\text{NN}}$ is a neural network with zero-mean i.i.d. parameters and continuous activation function $\phi$ which satisfies the linear envelope property in Equation \ref{eq:2}} and $g$ includes the following deterministic kernel-to-latent-space transformation for all $1 \leq i, j \leq N$
\begin{equation}
K_2 ( \boldsymbol{x}_i^{(2)}, \boldsymbol{x}_j^{(2)} ) \approx {\boldsymbol{z}_i^{(2)}}^T \boldsymbol{z}_j^{(2)} = g(\boldsymbol{x}_i^{(2)})^T g(\boldsymbol{x}_j^{(2)}), \label{eq:gapprox}
\end{equation}
where $K^{\text{NN}}_1$ is a NN-implied kernel and $K_2$ is any valid kernel of our choice.
\end{theorem}

To prove Theorem \ref{thm:1}, \textcolor{black}{we first state following lemma.}

\begin{lemma}
\label{lemma:2}
For latent representations $\boldsymbol{z}_i^{(1)}$ and $\boldsymbol{z}_j^{(1)}$ extracted from different data points $\boldsymbol{x}_i$ and $\boldsymbol{x}_j$ where $i \neq j$, the interactions between different entries of $\boldsymbol{z}_i^{(1)}$ and $\boldsymbol{z}_j^{(1)}$ can be reasonably ignored. In other words, let $\theta^{(1)}$ be the parameters of the neural network which takes in $\boldsymbol{x}^{(1)}$ and outputs $\boldsymbol{z}^{(1)}$, we have  $\mathbb{E}_{p(\theta^{(1)})} [ z_{ik}^{(1)}z_{jl}^{(1)} ] = 0$ for all $k \neq l$.
\end{lemma}

A detailed proof of Lemma \ref{lemma:2} is provided in Appendix \ref{appx:A}. With Lemma \ref{lemma:2}, let $\boldsymbol{\Theta} = \left\{\boldsymbol{\theta}^{(1)}, \boldsymbol{\theta}^{(2)}\right\}$ represent the parameters of ICK, we can calculate the covariance between $\hat{y}_i$ and $\hat{y}_j$ for different data indices $i \neq j$:
\begin{align}
&\text{cov}(\hat{y}_i, \hat{y}_j) \nonumber \\
&= \mathbb{E}_{p(\boldsymbol{\Theta})}[\hat{y}_i \hat{y}_j] - \mathbb{E}_{p(\boldsymbol{\Theta})}[\hat{y}_i] \mathbb{E}_{p(\boldsymbol{\Theta})}[\hat{y}_j] \label{eq:5} \\
&= \textstyle  \mathbb{E}_{p(\boldsymbol{\Theta})} \left[ \left(\sum_{k=1}^{p} z_{ik}^{(1)}z_{ik}^{(2)}\right) \left(\sum_{k=1}^{p} z_{jk}^{(1)}z_{jk}^{(2)}\right) \right] \label{eq:6} \\
&= \textstyle  \mathbb{E}_{p(\boldsymbol{\Theta})} \left[ \sum_{k=1}^{p} \sum_{l=1}^{p} z_{ik}^{(1)}z_{jl}^{(1)}z_{ik}^{(2)}z_{jl}^{(2)} \right] \label{eq:7} \\
&= \textstyle \mathbb{E}_{p(\boldsymbol{\Theta})} \left[ \sum_{k=1}^{p} z_{ik}^{(1)}z_{jk}^{(1)}z_{ik}^{(2)}z_{jk}^{(2)} \right] \label{eq:8} \\
&= \textstyle \sum_{k=1}^{p} \mathbb{E}_{p\left(\boldsymbol{\theta}^{(1)}\right)}\left[ z_{ik}^{(1)}z_{jk}^{(1)} \right] \mathbb{E}_{p\left(\boldsymbol{\theta}^{(2)}\right)}\left[ z_{ik}^{(2)}z_{jk}^{(2)} \right] \label{eq:9} \\
&\approx \textstyle K^{\text{NN}}_1 \left( \boldsymbol{x}_i^{(1)}, \boldsymbol{x}_j^{(1)} \right) \sum_{k=1}^{p} \mathbb{E}_{p\left(\boldsymbol{\theta}^{(2)}\right)}\left[ z_{ik}^{(2)}z_{jk}^{(2)} \right].
\label{eq:10}
\end{align}
Here, from Equation \ref{eq:5} to Equation \ref{eq:6}, we use the statement $\mathbb{E}_{p(\boldsymbol{\theta}^{(1)})}[z_{ik}^{(1)}] = 0$ in Section \ref{sec:3.3} and $\boldsymbol{\theta}^{(1)} \independent \boldsymbol{\theta}^{(2)}$, which leads to $\mathbb{E}_{p(\boldsymbol{\Theta})}[\hat{y}_i] = \mathbb{E}_{p(\boldsymbol{\Theta})}[\hat{y}_j] = 0$. From Equation \ref{eq:7} to Equation \ref{eq:8}, we get rid of all the cross terms using Lemma \ref{lemma:2} and $\boldsymbol{\theta}^{(1)} \independent \boldsymbol{\theta}^{(2)}$. \textcolor{black}{Specifically, we have $\mathbb{E}_{p(\boldsymbol{\Theta})} [ z_{ik}^{(1)}z_{jl}^{(1)}z_{ik}^{(2)}z_{jl}^{(2)} ] = \mathbb{E}_{p(\boldsymbol{\theta}^{(1)})} [ z_{ik}^{(1)}z_{jl}^{(1)} ] \mathbb{E}_{p(\boldsymbol{\theta}^{(2)})} [ z_{ik}^{(2)}z_{jl}^{(2)} ] = 0$ for all $k \neq l$.} From Equation \ref{eq:8} to Equation \ref{eq:9}, we again make use of $\boldsymbol{\theta}^{(1)} \independent \boldsymbol{\theta}^{(2)}$. From Equation \ref{eq:9} to Equation \ref{eq:10}, we use the statement $\mathbb{E}_{p(\boldsymbol{\theta}^{(1)})}[ z_{ik}^{(1)}z_{jk}^{(1)} ] \approx K^{\text{NN}} ( \boldsymbol{x}_i^{(1)}, \boldsymbol{x}_j^{(1)} )$ from Section \ref{sec:3.3}. If the kernel-to-latent-space transformation in $g(\cdot)$ is \emph{deterministic}, we can remove the expectation sign from the summation term in Equation \ref{eq:10} and the covariance can be further expressed as
\begin{equation}
\begin{split}
\text{cov}(\hat{y}_i, \hat{y}_j) &\approx K^{\text{NN}}_1 ( \boldsymbol{x}_i^{(1)}, \boldsymbol{x}_j^{(1)} ) ( {\boldsymbol{z}_i^{(2)}}^T \boldsymbol{z}_j^{(2)} ) \\
&= K^{\text{NN}}_1( \boldsymbol{x}_i^{(1)}, \boldsymbol{x}_j^{(1)} ) K_2 ( \boldsymbol{x}_i^{(2)}, \boldsymbol{x}_j^{(2)} ),\label{eq:final}  
\end{split}
\end{equation}
which means that $\hat{y}$ approximately follows a GP with a multiplicative composite kernel $K^{\text{comp}}(\boldsymbol{x}_i,\boldsymbol{x}_j) = K^{\text{NN}}_1\left(\boldsymbol{x}_i^{(1)},\boldsymbol{x}_j^{(1)}\right)K_2\left(\boldsymbol{x}_i^{(2)},\boldsymbol{x}_j^{(2)}\right)$ \textit{a priori}. This serves as a consise proof of Theorem 1. We also give a more detailed proof in Appendix \ref{appx:B}.

\subsection{Kernel-to-latent-space Transformation}
\label{sec:4.2}

We now show how we can construct an appropriate mapping $g(\cdot)$ that approximately satisfies the assumed  form of \eqref{eq:gapprox} and is used in the derivation of ICK from \eqref{eq:10} to \eqref{eq:final}. Here we adopt two methods, Nystr\"{o}m approximation and Random Fourier Features (RFF), to map the kernel matrix into the latent space. \textcolor{black}{Below, we elaborate the formulations for both methods, and give the methods and results for the Nystr\"{o}m method. The results of RFF will be presented in Appendix \ref{appx:C}.} We name our framework with N\emph{y}str\"{o}m method and \emph{r}andom Fourier Features ICK\emph{y} and ICK\emph{r}, respectively.

\begin{wrapfigure}{R}{0.5\textwidth}
\centering
\includegraphics[width=0.5\textwidth]{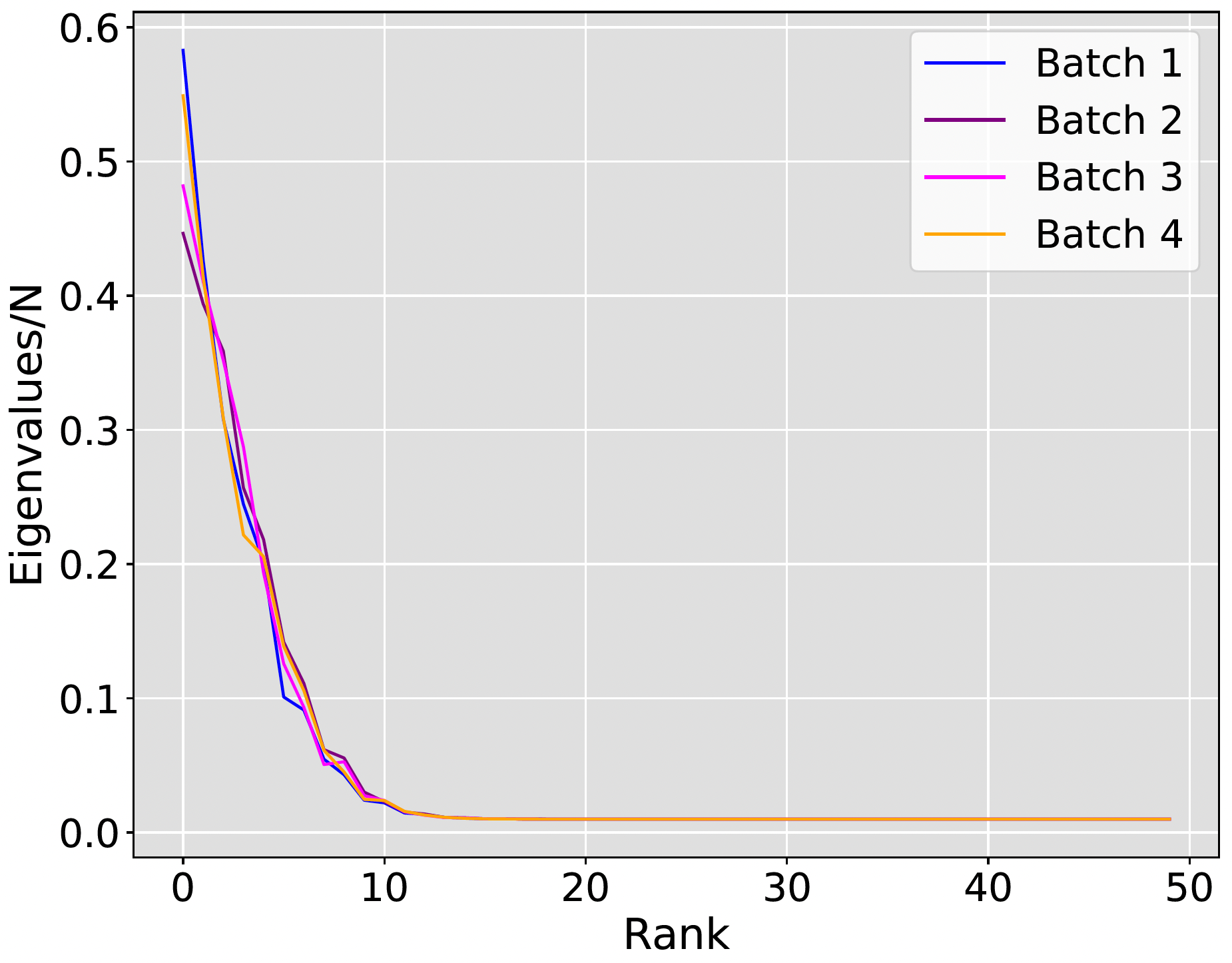}
\caption{Eigenvalues of the kernel matrix computed from the first 4 batches of training data where $N$ is the total number of data points.}
\label{fig:3}
\vspace{-4mm}
\end{wrapfigure}

\textcolor{black}{According to \citet{yang2012nystrom}, the Nystr\"{o}m method will yield much better performance than RFF if there exists a large gap in the eigen-spectrum of the kernel matrix. This phenomenon is mainly caused by how these two methods construct their basis functions. In particular, the basis functions used by RFF are sampled from a Gaussian distribution that is independent from the training examples, while the basis functions used by the Nystr\"{o}m method are sampled from the training samples so they are data-dependent. In our synthetic data experiments, we train our ICK framework using a batch size of 50. The eigenvalues of the kernel matrices computed from the first 4 batches of the synthetic data set are displayed in Figure \ref{fig:3}. It can be observed that the first few eigenvalues of the kernel matrix are much larger than the remaining eigenvalues. Namely, we observe a large gap in the eigen-spectrum of the kernel matrix, and Nystr\"{o}m method does generalize much better than RFF in our experiments.}

\subsubsection{Nystr\"{o}m Approximation}
\label{sec:4.2.1}

The main idea of Nystr\"{o}m approximation \citep{williams2000using} is to approximate the kernel matrix $\boldsymbol{K} \in \mathbb{R}^{N \times N}$ with a much smaller low-rank matrix $\boldsymbol{K}_q \in \mathbb{R}^{q \times q}$ where $q \ll N$ so both the computational and space complexity of kernel learning can be significantly reduced, yielding
\begin{equation}
\boldsymbol{K} \approx \hat{\boldsymbol{K}} = \boldsymbol{K}_{nq} \boldsymbol{K}_q^{-1} \boldsymbol{K}_{nq}^T.
\end{equation}
The entries of $\boldsymbol{K}_q$ and $\boldsymbol{K}_{nq}$ can be calculated as $\left(\boldsymbol{K}_q\right)_{ij} = K(\hat{x}_i, \hat{x}_j), i, j \in \{1, 2, ..., q\}$ and $\left(\boldsymbol{K}_{nq}\right)_{ij} = K(x_i, \hat{x}_j), i\in\{1, 2, ..., N\}, j\in\{1, 2, ..., q\}$, respectively. $x$ represents the original data points and $\hat{x}$ represents pre-defined inducing points (or pseudo-inputs \citep{snelson2005sparse}). In our study, these inducing points are chosen by defining an evenly spaced vector over the range of original data points. By performing Cholesky decomposition $\boldsymbol{K}_q^{-1} = \boldsymbol{U}^T \boldsymbol{U}$, where $\boldsymbol{U} \in \mathbb{R}^{q \times q}$, $\hat{\boldsymbol{K}}$ is expressed as
\begin{equation}
\begin{split}
\textstyle
\hat{\boldsymbol{K}} &= \boldsymbol{K}_{nq} \boldsymbol{K}_q^{-1} \boldsymbol{K}_{nq}^T \\ 
&= \boldsymbol{K}_{nq}\boldsymbol{U}^T \boldsymbol{U}\boldsymbol{K}_{nq}^T 
= (\boldsymbol{U}\boldsymbol{K}_{nq}^T)^T (\boldsymbol{U}\boldsymbol{K}_{nq}^T).
\end{split}
\end{equation}
Therefore, if we set $q = p$, then we can use $\boldsymbol{z}_i \triangleq \boldsymbol{U}(\boldsymbol{K}_{np}^T)_{:, i}$ as a kernel-to-latent-space transformation because each element in $\boldsymbol{K}$ approximately satisfies \eqref{eq:gapprox} as stated in Theorem 1: $K\left(x_i, x_j\right) = K_{ij} \approx \hat{K}_{ij} = \boldsymbol{z}_i^T \boldsymbol{z}_j$.  Conveniently, modern deep learning frameworks can propagate gradients through the Cholesky operation, making it straightforward to update the kernel parameters with gradient methods.  Note that as we increase the number of inducing points $p$, the approximation error between $\boldsymbol{K}$ and $\hat{\boldsymbol{K}}$ decreases. However, it is not recommended to set $p$ very large as updating the Cholesky decomposition requires $\mathcal{O}(p^3)$ . The empirical impact of $p$ on computational time and performance is shown in Appendix \ref{appx:D}.  In our experiments, only mild values of $p$ are necessary and the impact on computational is relatively small.  The full training procedure of ICK\emph{y} with $M = 2$ is presented in Algorithm \ref{alg:1}.

\subsubsection{Random Fourier Features}
\label{sec:4.2.2}

Random Fourier Features (RFF) is another popular approximation method used for kernel learning \citep{rahimi2007random}. Unlike the Nystr\"{o}m method which approximates the entire kernel matrix, RFF directly approximates the kernel function $K$ using some randomized feature mapping $\phi: \mathbb{R}^{D_m} \rightarrow \mathbb{R}^{2d_m}$ such that $K \left(x_i^{(m)}, x_j^{(m)}\right) \approx \phi\left(x_i^{(m)}\right)^T \phi\left(x_j^{(m)}\right)$. To obtain the feature mapping $\phi$, based on Bochner's theorem, we first compute the Fourier transform $p(\omega)$ of kernel $K$
\begin{equation}
p(\omega) = \frac{1}{(2\pi)^{D_m}} \int_{-\infty}^{+\infty} e^{-j\omega^T \delta} K(\delta)d\delta, 
\end{equation}
where $\delta = x_i^{(m)} - x_j^{(m)}$. Then we draw $d_m$ i.i.d. samples $\omega_1, \omega_2, ..., \omega_{d_m}$ from $p(\omega)$ and construct the feature mapping $\phi$ as follows
\begin{dmath}
\phi \left(x^{(m)}\right)
\equiv \textstyle d_m^{-1/2}\left[ \cos\left(\omega_1^T x^{(m)}\right), ..., \cos\left(\omega_{d_m}^T x^{(m)}\right),
\sin\left(\omega_1^T x^{(m)}\right),
..., \sin\left(\omega_{d_m}^T x^{(m)}\right) \right].
\end{dmath}
Since $\phi \left(x^{(m)}\right) \in \mathbb{R}^{2d_m}$, we need to set $d_m = p/2$ when using RFF as a kernel-to-latent-space transformation. In addition, since RFF involves sampling from a distribution, the kernel parameters are thus not directly differentiable and we need to apply a reparameterization trick \citep{maddison2016concrete} to learn those parameters. The full training procedure of ICK\emph{r} with $M = 2$ is presented in Algorithm \ref{alg:2}.

\subsection{Uncertainty Estimation}
\label{sec:4.3}

\begin{wrapfigure}{R}{0.5\textwidth}
\vspace{-6mm}
\includegraphics[width=0.5\textwidth]{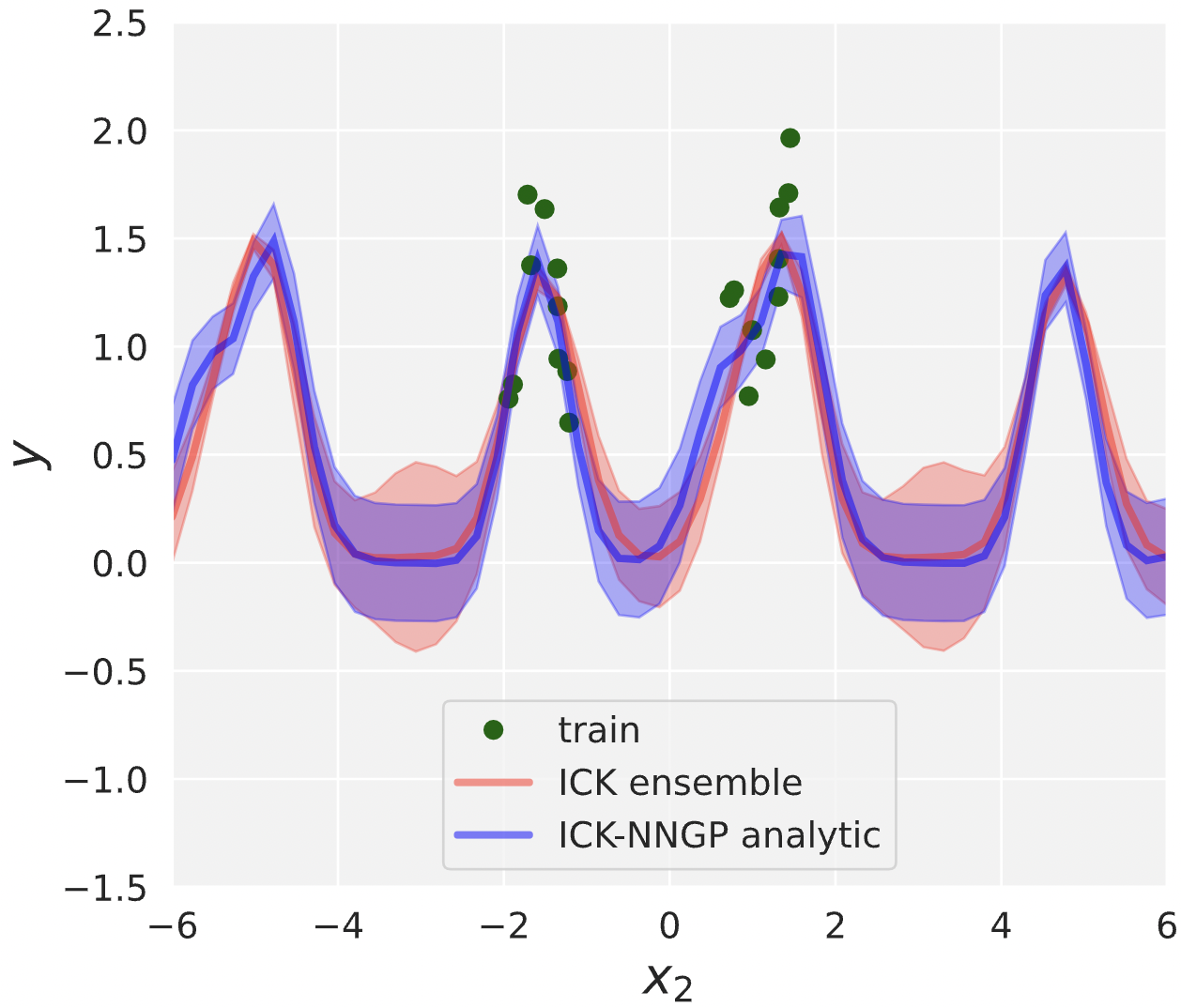}
\vspace{-6mm}
\caption{Predictive distribution from ICK\emph{y} ensemble and its GP posterior counterpart on a 1D regression task.}
\label{fig:4}
\end{wrapfigure}

\textcolor{black}{Since the motivation of ICK discussed in Section \ref{sec:4.1} is closely related to GPs, we present two distinct approaches for estimating uncertainty in the subsequent sections: ensembling method and direct calculation of posterior variance.}

\begin{figure}[t!]
\centering
\begin{algorithm}[H]
\caption{Implicit Composite Kernel-Nystr\"{o}m (ICK\emph{y})} \label{alg:1}
\textbf{Input:} data $\boldsymbol{X} = \left\{\boldsymbol{x}_i^{(1)}, \boldsymbol{x}_i^{(2)}\right\}_{i=1}^N$, targets $\boldsymbol{y} = [y_i]_{i=1}^N$, $f_{\text{NN}}, \boldsymbol{\theta}^{(1)}, K^{(2)}, \boldsymbol{\theta}^{(2)}$, learning rate $\beta$
\begin{algorithmic}
\State Sample a total of $N_B$ minibatches $\{\boldsymbol{X}_B, \boldsymbol{y}_B\}_{B=1}^{N_B}$
\For{$B$ from 1 to $N_B$}
\For{$\boldsymbol{x}_i^{(1)}, \boldsymbol{x}_i^{(2)}$ in $\boldsymbol{X}_B, i = 1,..., n_B$}
\State $\boldsymbol{z}_i^{(1)} = f_{\text{NN}} \left(\boldsymbol{x}_i^{(1)}\right)$
\State Define inducing points $\hat{\boldsymbol{x}}_1^{(2)}, ..., \hat{\boldsymbol{x}}_p^{(2)}$
\State Compute $\boldsymbol{K}_p$: $\left( \boldsymbol{K}_p \right)_{jk} = K^{(2)} \left( \hat{\boldsymbol{x}}_j^{(2)}, \hat{\boldsymbol{x}}_k^{(2)} \right)$
\State Do Cholesky decomposition $\left(\boldsymbol{K}_p\right)^{-1} = \boldsymbol{U}^T \boldsymbol{U}$
\State Compute $\boldsymbol{K}_{np}$: $\left(\boldsymbol{K}_{np}\right)_{jk} = K^{(2)}\left(\boldsymbol{x}_j^{(2)}, \hat{\boldsymbol{x}}_k^{(2)}\right)$
\State $\boldsymbol{z}_i^{(2)} = \boldsymbol{U} \left(\boldsymbol{K}_{np}^T\right)_{:,i}$
\State $\hat{y}_i = {\boldsymbol{z}_i^{(1)}}^T \boldsymbol{z}_i^{(2)}$
\EndFor
\State $\hat{\boldsymbol{y}}_B = \texttt{concat}\left(\hat{y}_1, ..., \hat{y}_{n_B}\right)$
\State Compute loss $\mathcal{L} = \mathcal{L}\left(\boldsymbol{y}_B, \hat{\boldsymbol{y}}_B\right)$
\State $\boldsymbol{\theta}^{(1)} \leftarrow \boldsymbol{\theta}^{(1)} - \beta \nabla_{\boldsymbol{\theta}^{(1)}} \mathcal{L}$
\State $\boldsymbol{\theta}^{(2)} \leftarrow \boldsymbol{\theta}^{(2)} - \beta \nabla_{\boldsymbol{\theta}^{(2)}} \mathcal{L}$
\EndFor
\end{algorithmic}
\textbf{Return:} Predictions $\hat{\boldsymbol{y}}$ and updated parameters $\boldsymbol{\theta}^{(1)}, \boldsymbol{\theta}^{(2)}$
\end{algorithm}
\vspace{-6mm}
\end{figure}

\subsubsection{Ensembling}
\label{sec:4.3.1}

One approach for estimating the uncertainty is to adopt a sample-then-optimize approach \citep{matthews2017sample} and construct a deep ensemble posterior approximation for ICK. \textcolor{black}{There are several ways to enable GP posterior interpretation for a deep ensemble trained with SGD.} Specifically, let the final layer of each baselearner NN be $C$-dimensional and denote the deep ensemble as $F = \{f_{n_e}\}_{n_e = 1}^{N_e}$, where $f_{n_e}$ is a baselearner NN and $N_e$ is the total number of baselearners in the ensemble. After training, each $f_{n_e}$ can be viewed as i.i.d. samples from a multi-output GP posterior with kernel function $K_F$ in the infinite width limit and the ensemble $F$ thus represents $N_e$ independent draws from the GP posterior, where $K_F$ can be either the \emph{Neural Network Gaussian Process} (NNGP) kernel \citep{lee2017deep} \textcolor{black}{if we make only the last layer trainable} or the \emph{Neural Tangent Kernel} (NTK) \citep{he2020bayesian,jacot2018neural} \textcolor{black}{if we add a randomized and untrainable function to each baselearner.}

\begin{figure}[t!]
\centering
\begin{algorithm}[H]
\caption{Implicit Composite Kernel-RFF (ICK\emph{r})} \label{alg:2}
\textbf{Input:} data $\boldsymbol{X} = \left\{\boldsymbol{x}_i^{(1)}, \boldsymbol{x}_i^{(2)}\right\}_{i=1}^N$, targets $\boldsymbol{y} = [y_i]_{i=1}^N$, $f_{\text{NN}}, \boldsymbol{\theta}^{(1)}, K_2, \boldsymbol{\theta}^{(2)}$, learning rate $\beta$
\begin{algorithmic}
\State Sample a total of $N_B$ minibatches $\{\boldsymbol{X}_B, \boldsymbol{y}_B\}_{B=1}^{N_B}$
\For{$B$ from 1 to $N_B$}
\For{$\boldsymbol{x}_i^{(1)}, \boldsymbol{x}_i^{(2)}$ in $\boldsymbol{X}_B, i = 1,..., n_B$}
\State $\boldsymbol{z}_i^{(1)} = f_{\text{NN}} \left(\boldsymbol{x}_i^{(1)}\right)$
\textcolor{black}{
\State Compute the Fourier transform $p(\omega)$ of kernel $K_2$:
\State $p(\omega) = \frac{1}{(2\pi)^{D_m}} \int_{-\infty}^{+\infty} e^{-j\omega^T \delta} K(\delta)d\delta$
\State where $\delta = x_i^{(2)} - x_j^{(2)}$
\State Draw $p/2$ i.i.d. samples $\omega_1, \omega_2, ..., \omega_{p/2}$ from $p(\omega)$.
\State Construct feature mapping 
\State $\boldsymbol{z}_i^{(2)} = \sqrt{p/2} \Big[ \cos \big(\omega_1^T x^{(2)}\big), ..., \cos\big(\omega_{p/2}^T x^{(2)}\big),$
\State $\sin\big(\omega_1^T x^{(2)}\big),
..., \sin\big(\omega_{p/2}^T x^{(2)}\big) \Big]$
}
\State $\hat{y}_i = {\boldsymbol{z}_i^{(1)}}^T \boldsymbol{z}_i^{(2)}$
\EndFor
\State $\hat{\boldsymbol{y}}_B = \texttt{concat}\left(\hat{y}_1, ..., \hat{y}_{n_B}\right)$
\State Compute loss $\mathcal{L} = \mathcal{L}\left(\boldsymbol{y}_B, \hat{\boldsymbol{y}}_B\right)$
\State $\boldsymbol{\theta}^{(1)} \leftarrow \boldsymbol{\theta}^{(1)} - \beta \nabla_{\boldsymbol{\theta}^{(1)}} \mathcal{L}$
\State $\boldsymbol{\theta}^{(2)} \leftarrow \boldsymbol{\theta}^{(2)} - \beta \nabla_{\boldsymbol{\theta}^{(2)}} \mathcal{L}$
\EndFor
\end{algorithmic}
\textbf{Return:} Predictions $\hat{\boldsymbol{y}}$ and updated parameters $\boldsymbol{\theta}^{(1)}, \boldsymbol{\theta}^{(2)}$
\end{algorithm}
\vspace{-6mm}
\end{figure}

\begin{figure}[t!]
\centering
\begin{algorithm}[H]
\caption{ICK\emph{y} Ensemble} \label{alg:3}
\textbf{Input:} data $\boldsymbol{X} = \left\{\boldsymbol{x}_i^{(1)}, \boldsymbol{x}_i^{(2)}\right\}_{i=1}^N$, targets $\boldsymbol{y} = [y_i]_{i=1}^N$, ensemble $F_{\text{ICK}y} = \left\{f_1, ..., f_{N_e}\right\}$ where each function in $F$ consists of $f_{\text{NN}}, K_2$ with parameters $\boldsymbol{\theta}^{(1)}, \boldsymbol{\theta}^{(2)}$, respectively
\begin{algorithmic}
\For{$s = 1, ..., N_e$}
\State Apply proper initialization strategy to $f_s$
\State Perform Algorithm \ref{alg:1} on function $f_s$
\EndFor
\end{algorithmic}
\textbf{Return:} 
predictive mean $\hat{\boldsymbol{\mu}} = \frac{1}{N_e} \sum_{s=1}^{N_e} f_s(\boldsymbol{X})$, 
predictive variance $\hat{\boldsymbol{\sigma}}^2 = \frac{1}{N_e} \sum_{s=1}^{N_e} \left(f_s(\boldsymbol{X}) - \hat{\boldsymbol{\mu}}\right)^2$, and
updated parameters $\boldsymbol{\theta}^{(1)}, \boldsymbol{\theta}^{(2)}$ for all functions $f$ in $F_{\text{ICK}y}$
\end{algorithm}
\vspace{-6mm}
\end{figure}

This deep ensemble mechanism can be easily applied to ICK\emph{y}. Specifically, the final prediction of ICK\emph{y} $\hat{y} = f_{\text{ICK}y}\left(\boldsymbol{x}^{(1)}, \boldsymbol{x}^{(2)}\right)$ can be viewed as a weighted sum of  $\hat{y} = \sum_{k=1}^p \alpha_k z_k^{(1)}$ where $\alpha_k = z_k^{(2)} = g\left(\boldsymbol{x}^{(2)}\right)_k$. Hence, if we construct an ensemble of ICK\emph{y}, $F_{\text{ICK}y} = \left\{f_1, ..., f_{N_e}\right\}$ (as shown in Algorithm \ref{alg:3}), and $f_{\text{NN}}$ is appropriately initialized for all baselearners $f_s \in F_{\text{ICK}y}, s = 1, ..., N_e$,
then all trained baselearners in $F_{\text{ICK}y}$ can be \emph{approximately} viewed as i.i.d. posterior samples from a single-output GP with a multiplicative kernel $K_F K_2$. $K_F$ will either be an NNGP or an NTK and $K_2$ comes from our defined kernel. If the parameters of $f_{\text{NN}}$ (i.e. $\boldsymbol{\theta}^{(1)}$) are independently drawn from a zero-mean Gaussian and are fixed except the last layer (as specified by \citet{lee2019wide}) for all baselearners in the ensemble, this corresponds to an NNGP. A detailed proof is in Appendix \ref{appx:E}.

To verify our argument, we train an ICK\emph{y} ensemble $F_{\text{ICK}y}$ containing 300 baselearners on the same 1D regression task as provided by \citet{he2020bayesian} using SGD optimizer \textcolor{black}{with a learning rate of $5*10^{-4}$ and the mean-squared error (MSE) loss until convergence}. We then compare the predictive distribution to its GP posterior counterpart based on the NNGP implementation in the \emph{neural\_tangents} \citep{neuraltangents2020} package. \textcolor{black}{The weights for each baselearner are initialized to be drawn from an i.i.d. Gaussian with zero mean and variance equal to $\sigma_w / \sqrt{d_N}$ where $d_N$ is the width of the corresponding NN layer.} The GP posterior has zero mean and kernel function $K_{\text{NNGP}}K_2$, where $K_{\text{NNGP}}$ is the corresponding NNGP kernel of $f_{\text{NN}}$ and $K_2$ is the chosen kernel function for the mapping $g$ in the ICK\emph{y} ensemble in Figure \ref{fig:4}. Here, $K_2$ is set to be an \emph{exponential-sine-squared} kernel with period $T = 2\pi$. The predictive distribution of ICK\emph{y} ensemble is very close to the analytic GP posterior, demonstrating the \emph{approximate} equivalence between the two models.

\subsubsection{Computing Variance}
\label{sec:4.3.2}

\begin{figure}[t!]
\centering
\subfloat[\label{fig:5a}]{
\includegraphics[width=0.48\linewidth]{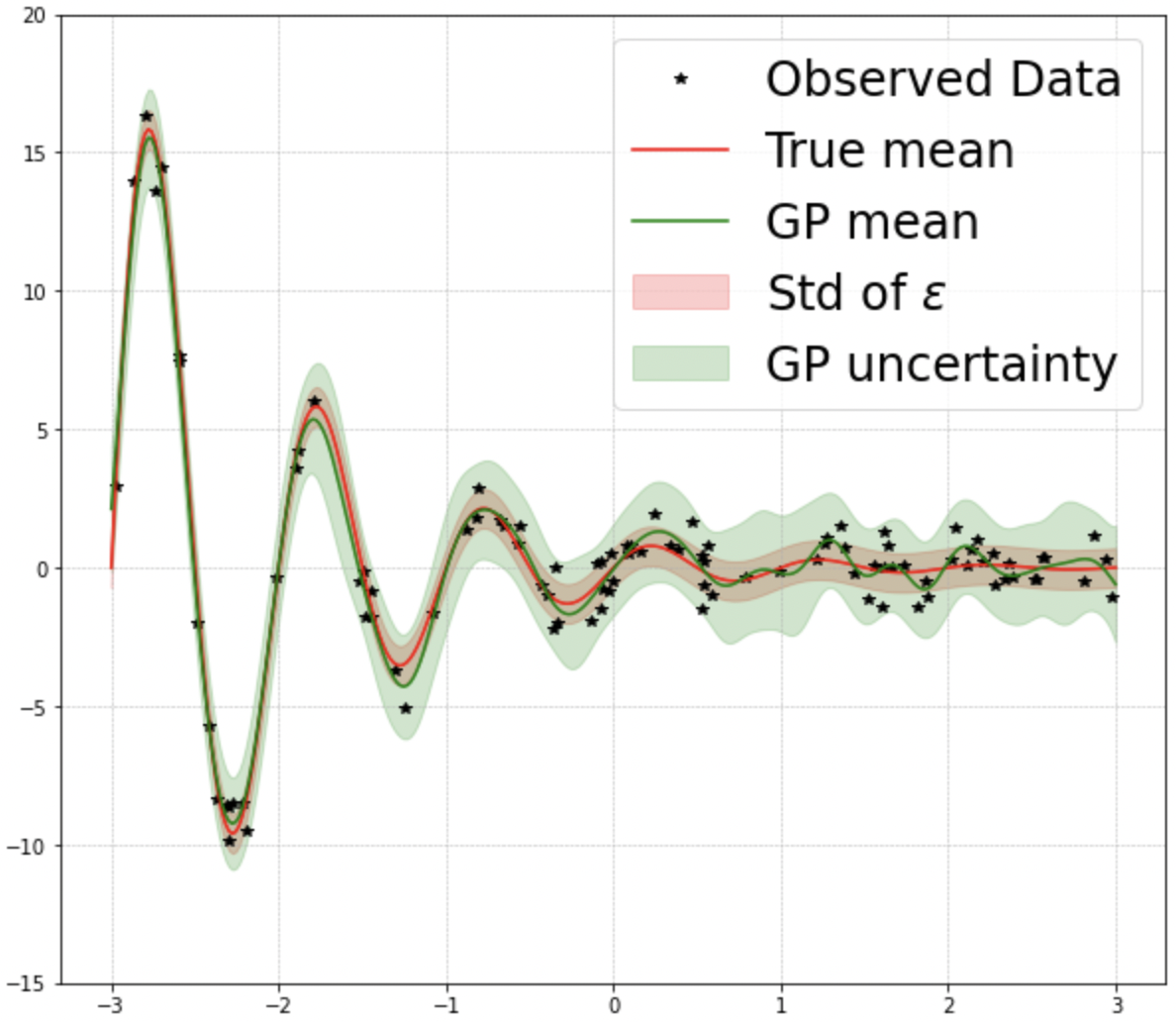}}
\hfill
\subfloat[\label{fig:5b}]{
\includegraphics[width=0.48\linewidth]{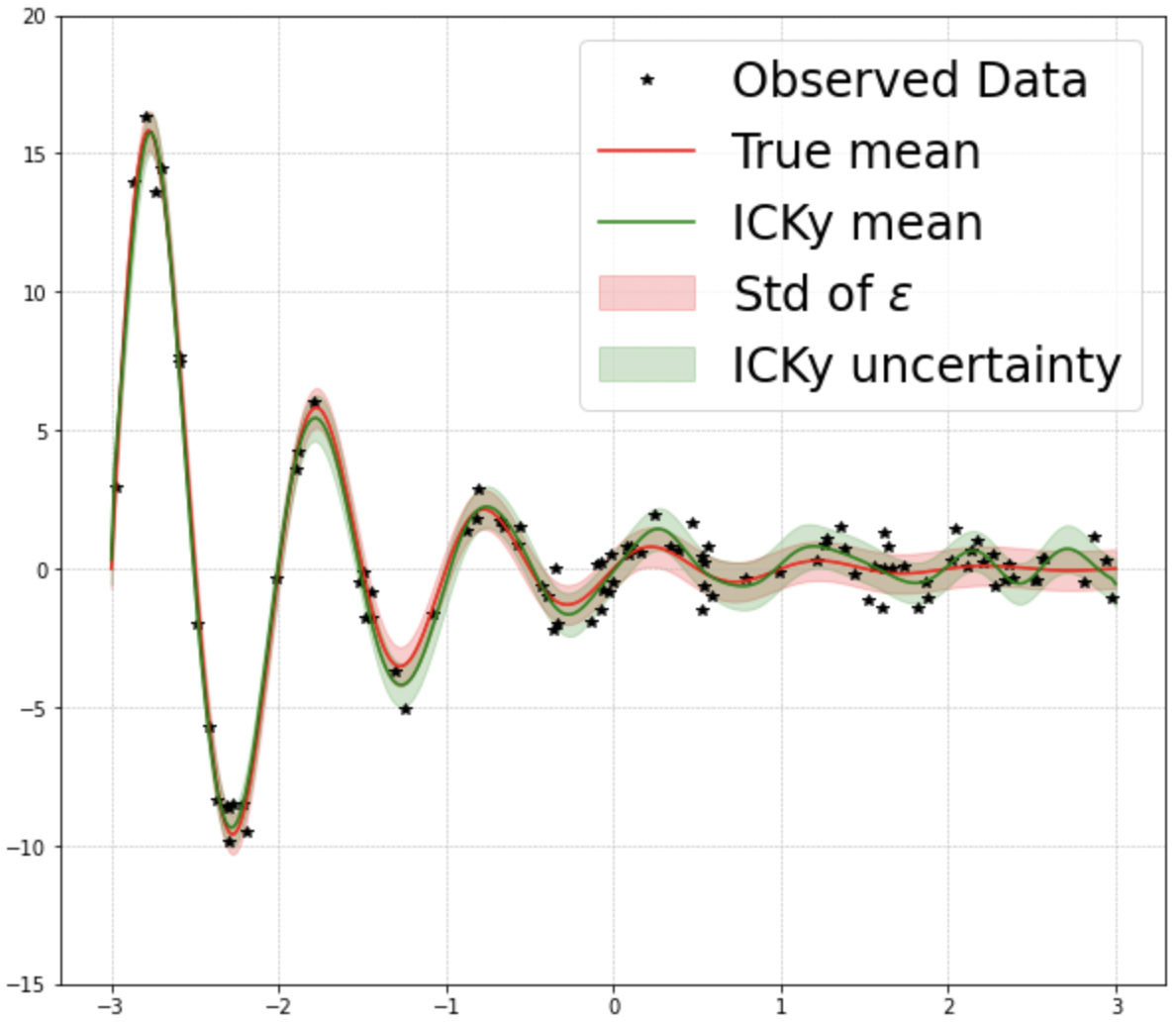}}
\caption{Plots of predicted mean and uncertainties for a sinusoidal function with exponential damping with noise $f(x) = e^{-x} \sin(2\pi x) + \epsilon$ using (a) an exact GP with RBF kernel and (b) ICK\emph{y} with RBF kernel. Here ICK\emph{y} predicts the uncertainty by directly calculating the covariance matrix $\boldsymbol{\Sigma}$ in Equation \ref{eq:17}.}
\label{fig:5}
\end{figure}

\textcolor{black}{Another approach to estimate the uncertainty is to directly compute the covariance matrix of the posterior distribution}
\begin{equation}
\boldsymbol{\Sigma} = \boldsymbol{K}_{x^{*}x^{*}} - \boldsymbol{K}_{x^{*}x} \left( \boldsymbol{K}_{xx} + \sigma^2 \boldsymbol{I} \right) \boldsymbol{K}_{xx^{*}}, 
\label{eq:17}
\end{equation}
where $\boldsymbol{K}_{x^{*}x^{*}} = K(\boldsymbol{X}^*, \boldsymbol{X}^*)$ is the kernel matrix evaluated on the test dataset, $\boldsymbol{K}_{xx} = K(\boldsymbol{X}, \boldsymbol{X})$ is the kernel matrix evaluated on the training dataset, $\boldsymbol{K}_{x^{*}x} = K(\boldsymbol{X}^*, \boldsymbol{X})$ is the cross-covariance matrix evaluated on both training and test datasets, and $\boldsymbol{K}_{x^{*}x} = \boldsymbol{K}_{xx^{*}}^{T}$. Here $K$ refers to the composite kernel function implied by ICK. Note that if we get rid of $f_{\text{NN}}(\cdot)$ and only keep $g(\cdot)$, then $K = K_2$ and the computation of $\boldsymbol{\Sigma}$ is straightforward. \textcolor{black}{We believe the true uncertainty can be reasonably approximated by ICK\emph{y}, as indicated by $\text{diag}(\boldsymbol{\Sigma})$, while maintaining a reasonable computational complexity.} To demonstrate this, we sample 100 data points from a sinusoidal function with exponential decay $f(x) = e^{-x} \sin(2\pi x) + \epsilon$ where $\epsilon \sim \mathcal{N}(0, 0.5)$ and see how an exact GP and ICK\emph{y} with radial basis function (RBF) kernel can capture the mean and uncertainty of this function by learning from these samples and directly calculating the covariance $\boldsymbol{\Sigma}$ according to Equation \ref{eq:17}. In Figure \ref{fig:5}, our results reveal that both GP and ICK\emph{y} accurately capture the mean function. However, ICK\emph{y} exhibits a prediction of uncertainty closer to the ground truth when compared to GP.

\section{Experiments}
\label{sec:5}

We evaluate ICK\emph{y} on \textcolor{black}{6 different data sets: 2 synthetic datasets and 4 real-world datasets \footnotemark{}}. In all the experiments, our ICK\emph{y} framework only consists of 2 kernels (i.e. $M = 2$), one NN-implied kernel and one chosen kernel function with trainable parameters. The implementation details of all the experiments in this section are provided in Appendix \ref{appx:F} and the data accessibility and restrictions are provided in Appendix \ref{appx:G}.
\footnotetext[1]{The code for all the experiments presented in this paper can be accessed at: \url{https://github.com/jzy95310/ICK}.}

\subsection{Synthetic Data}
\label{sec:5.1}

\subsubsection{Case when \texorpdfstring{$M = 2$}{M = 2}}
\label{sec:5.1.1}
To verify that ICK\emph{y} can simulate sampling from a GP with \emph{multiplicative kernel}, we create a synthetic data set $y \sim \mathcal{GP}(0, K_1 K_2)$ containing 3000 data points where $\boldsymbol{x}^{(1)} \in [0,1]$ is the input for the linear kernel $K_1$ and $\boldsymbol{x}^{(2)} \in [0,2]$ is the input for a \emph{spectral mixture} kernel \citep{wilson2013gaussian} $K_2$ with 2 components.  

\begin{figure}[t!]
\centering
\includegraphics[scale=0.33]{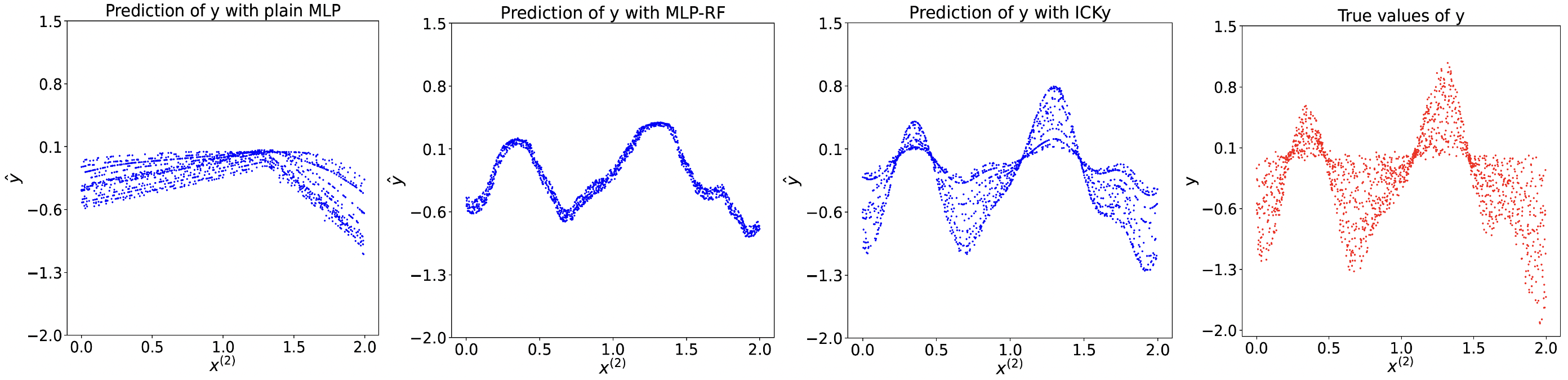}
\caption{Prediction of $y \sim \mathcal{GP}(0, K_1 K_2)$, where $\boldsymbol{x}^{(1)}$ is input to a linear kernel $K_1$ and $\boldsymbol{x}^{(2)}$ is input to a spectral mixture kernel $K_2$. We plot $\boldsymbol{x}^{(2)}$ against the predicted $y$. We show results from a plain MLP (left), MLP-RF (middle left), and ICK\emph{y} framework (middle right), and we compare to the true values of $y$ (right).}
\label{fig:6}
\end{figure}

\begin{figure}[t!]
\centering
\includegraphics[scale=0.33]{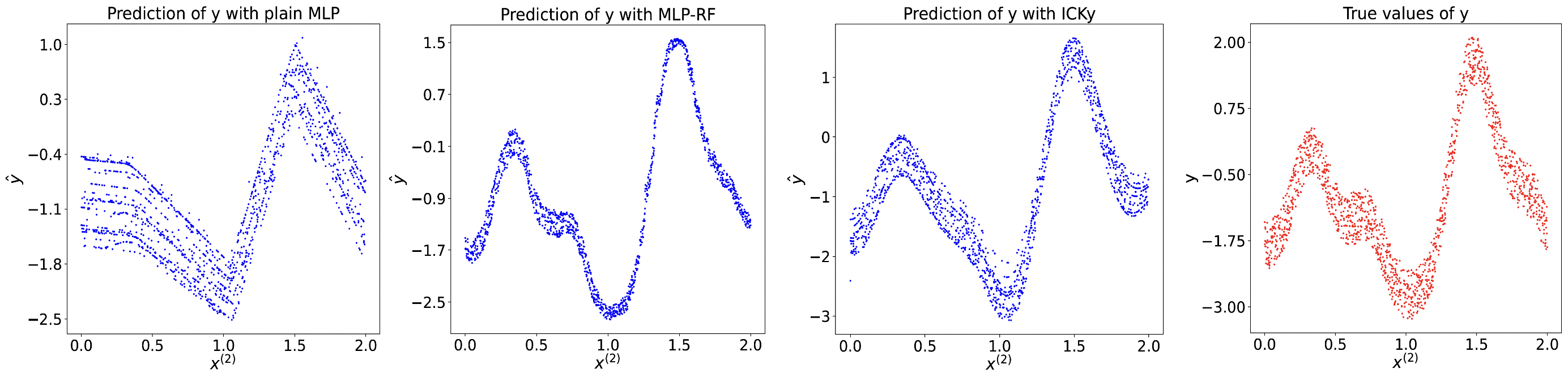}
\caption{Prediction of $y \sim \mathcal{GP}(0, K_1 + K_2)$, where $\boldsymbol{x}^{(1)}$ is input to a linear kernel $K_1$ and $\boldsymbol{x}^{(2)}$ is input to a spectral mixture kernel $K_2$. We plot $\boldsymbol{x}^{(2)}$ against the predicted $y$. We show results from a plain MLP (left), MLP-RF (middle left), and ICK\emph{y} framework (middle right), and we compare to the true values of $y$ (right).}
\label{fig:7}
\end{figure}

We compare ICK\emph{y} with two models: a multi-layer perceptron (MLP) applied to the concatenated features and a novel multi-layer perceptron-random forest (MLP-RF) joint model employed by \citet{zheng2021local}, where MLP learns from $\boldsymbol{x}^{(1)}$ and RF learns from $\boldsymbol{x}^{(2)}$. We believe MLP-RF serves as a good benchmark model as it is a joint model with similar architecture to our ICK\emph{y} framework. To see how ICK\emph{y} simulates the spectral mixture kernel, we plot only $\boldsymbol{x}^{(2)}$ against the predicted value of $y$ as shown in Figure \ref{fig:6}. As can be seen from the figure, plain MLP only captures the linear trend. MLP-RF only captures the mean of the spectral mixture components. In contrast, our ICK\emph{y} framework captures both the mean and the variance of the spectral mixture kernel. We also repeat this experiment by sampling from GP with an \emph{additive kernel} and we obtain similar observations as shown in Figure \ref{fig:7}.

\begin{figure}[t!]
\centering
\subfloat[\label{fig:8a}]{
\includegraphics[width=0.32\linewidth]{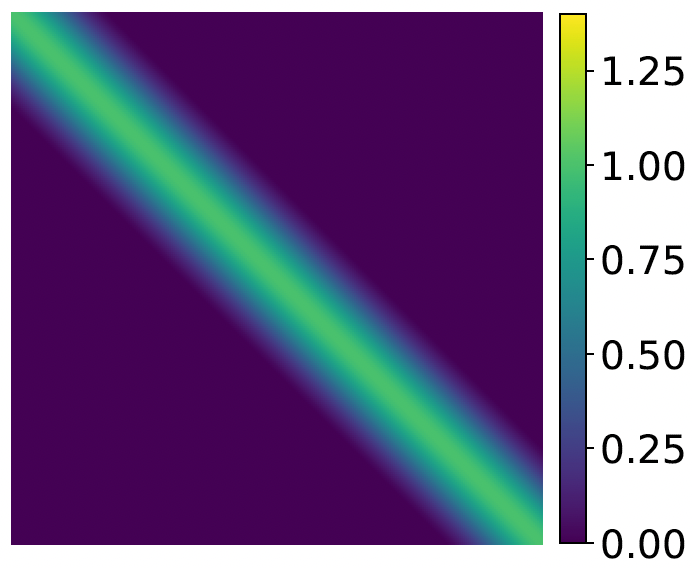}}
\subfloat[\label{fig:8b}]{
\includegraphics[width=0.32\linewidth]{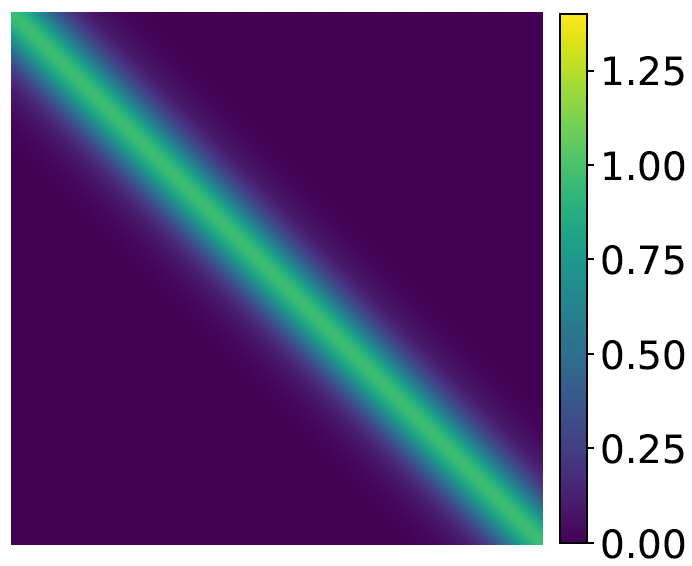}}
\subfloat[\label{fig:8c}]{
\includegraphics[width=0.32\linewidth]{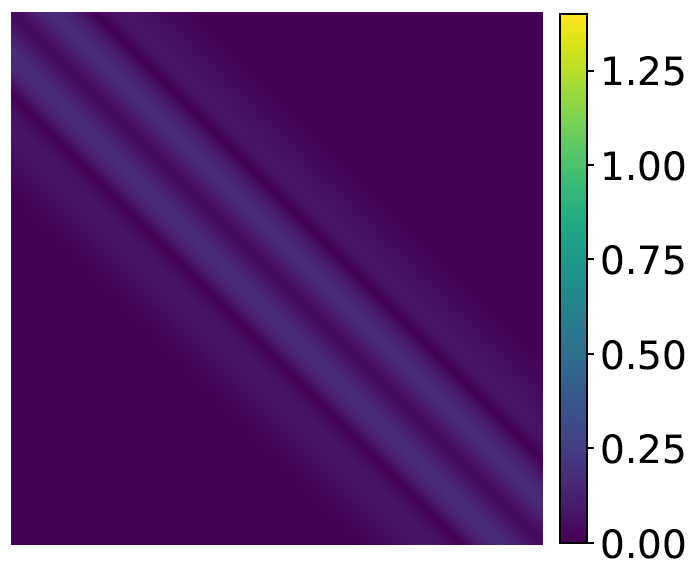}}
\caption{Visualization of (a) True matrix (b) estimated matrix by our ICK\emph{y} framework, and (c) absolute difference between the true and estimated matrix for the spectral mixture kernel}
\label{fig:8}
\end{figure}

\textcolor{black}{We also examine whether ICK\emph{y} can retrieve the spectral mixture kernel in this task. After fitting the parameters of the spectral mixture kernel in ICK\emph{y}, we compute the kernel matrix $\boldsymbol{K}_{\text{ICK\emph{y}}}$ using these learned parameters and compare it with the true kernel matrix $\boldsymbol{K}_{\text{true}}$ by calculating the absolute difference between them as displayed in Figure \ref{fig:8}. As can be observed, $\boldsymbol{K}_{\text{ICK\emph{y}}}$ and $\boldsymbol{K}_{\text{true}}$ are similar and their absolute difference is relatively small, indicating that ICK\emph{y} can approximately retrieve the spectral mixture kernel.}

\begin{figure}[t!]
\centering
\subfloat[\label{fig:9a}]{
\includegraphics[width=0.32\linewidth]{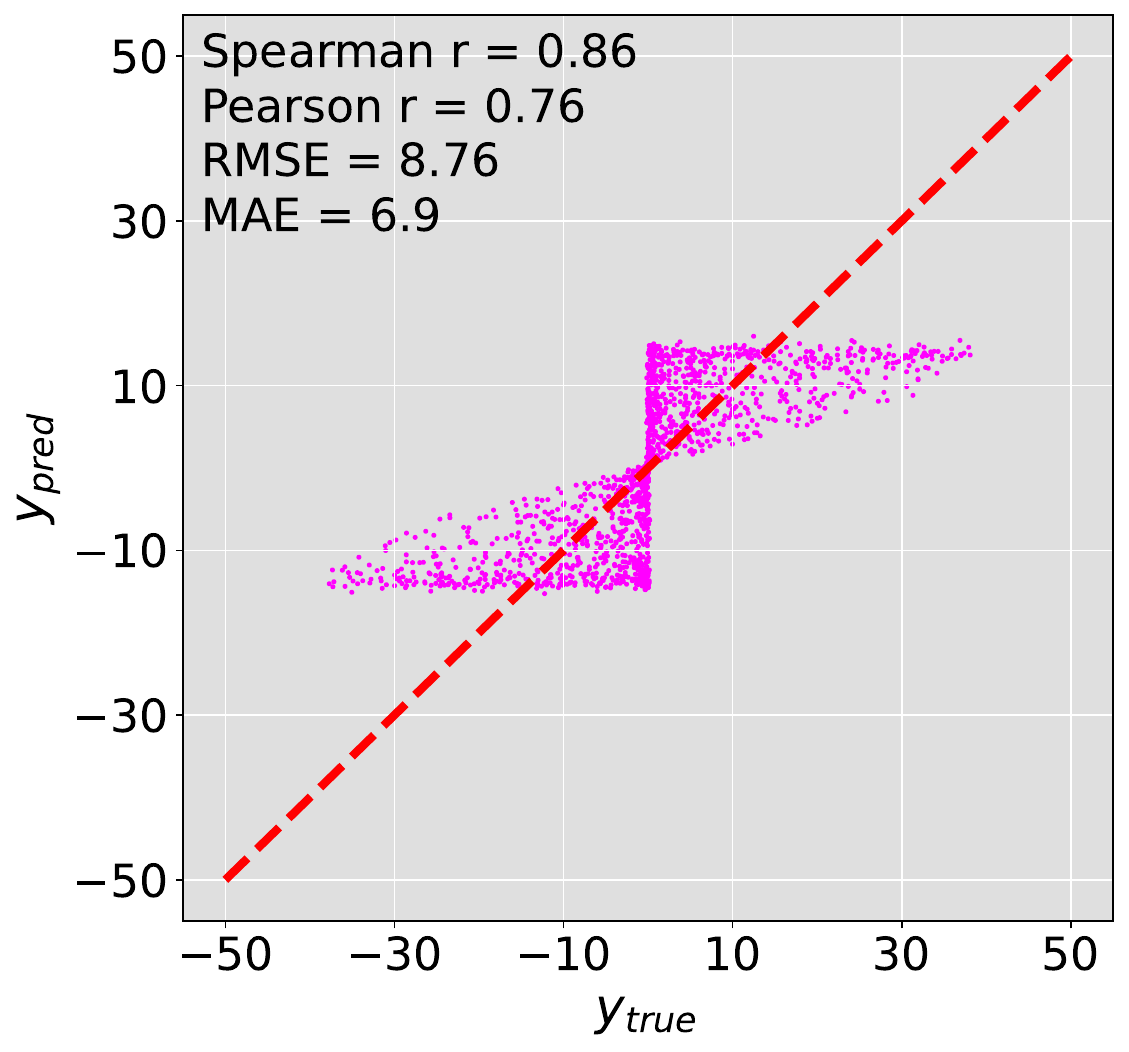}}
\subfloat[\label{fig:9b}]{
\includegraphics[width=0.32\linewidth]{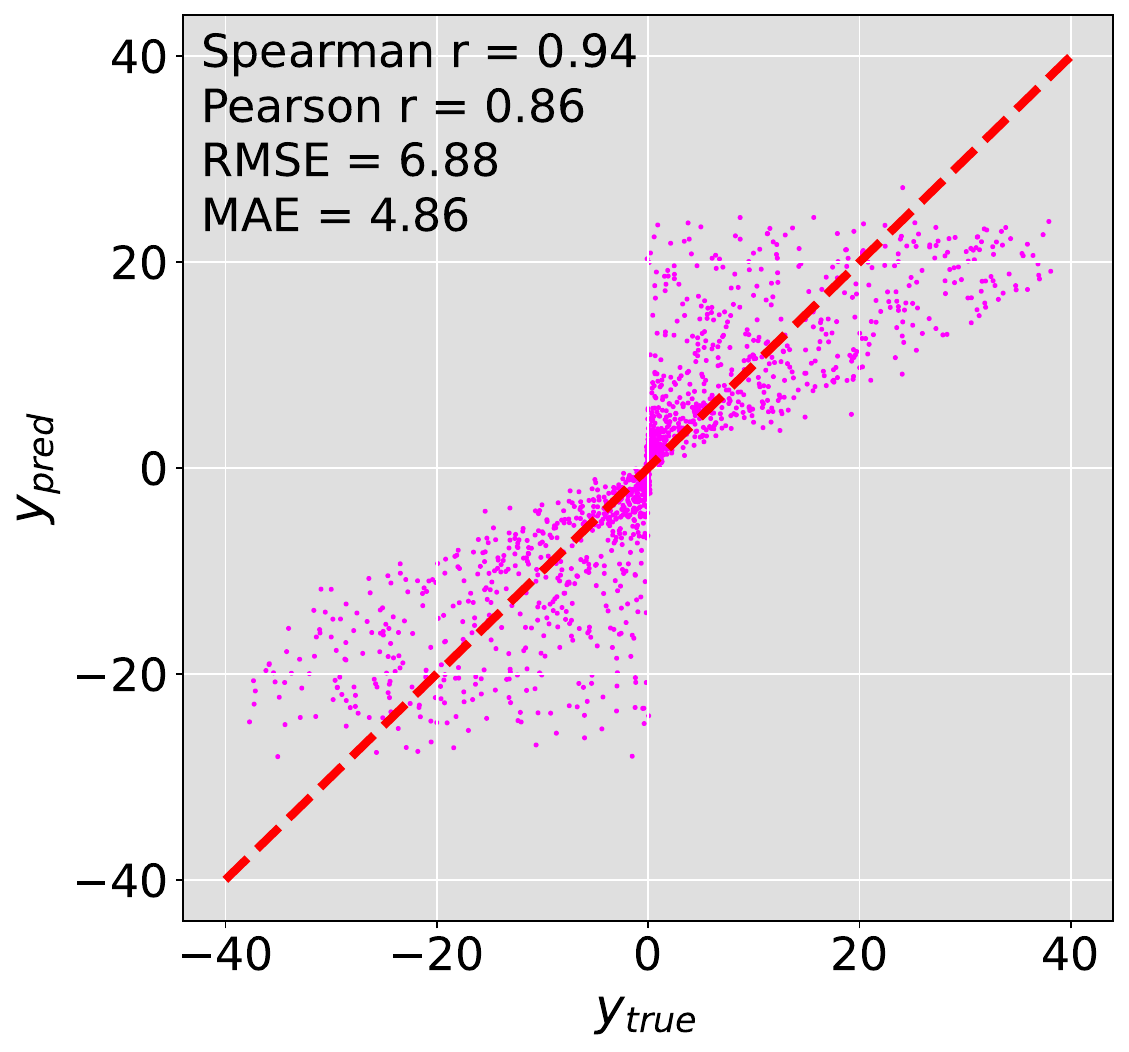}}
\subfloat[\label{fig:9c}]{
\includegraphics[width=0.32\linewidth]{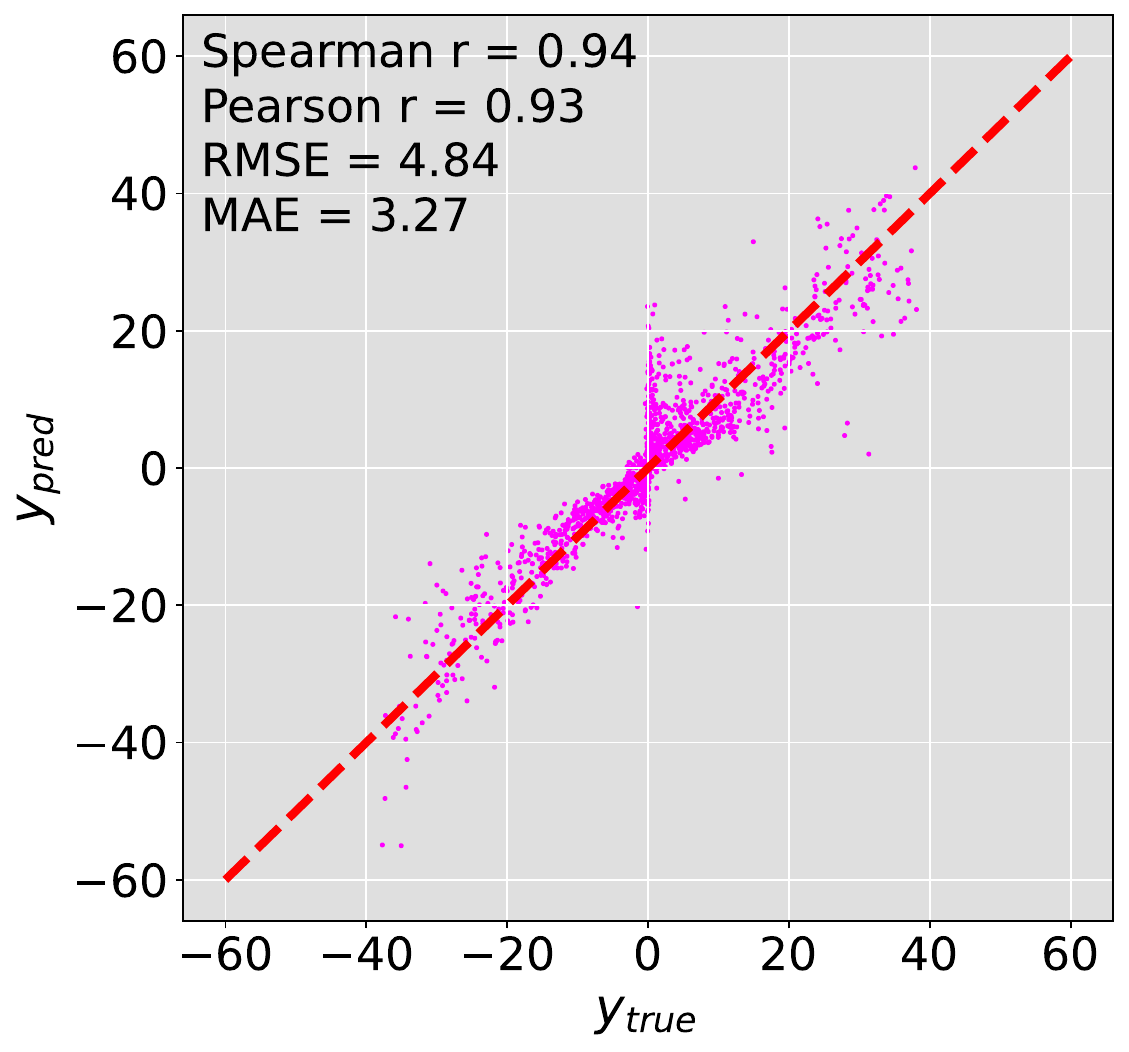}}
\caption{Scatter plots of the true values of $y$ against the predicted values of $y$ using our ICK\emph{y} framework with information from (a) one source $\hat{y} = f_{\text{NN}}\left(x^{(1)}\right)$, (b) 2 sources $\hat{y} = f_{\text{ICK\emph{y}}}\left(x^{(1)}, x^{(2)}\right)$, and (c) 3 sources $\hat{y} = f_{\text{ICK\emph{y}}}\left(x^{(1)}, x^{(2)}, x^{(3)}\right)$.}
\label{fig:9}
\end{figure}

\subsubsection{Case when \texorpdfstring{$M = 3$}{M = 3}}
\label{sec:5.1.2}

\textcolor{black}{We then construct another synthetic data set with 3000 data points where each input $\boldsymbol{x} = \left\{ x^{(1)}, x^{(2)}, x^{(3)} \right\}$ contains information from 3 difference sources. The output $y$ is generated by $y = x^{(3)}\tanh{\left(2x^{(1)} \cos^2\left(\pi x^{(2)}/50\right)\right)} + \epsilon$ where $\epsilon$ is a Gaussian noise term. We process $x^{(1)}$ with a small single-hidden-layer NN, $x^{(2)}$ with an \emph{exponential sine squared} kernel, and $x^{(3)}$ with a \emph{radial-basis function} (RBF) kernel. Figure \ref{fig:9} shows the prediction results as we progressively add more sources of information into our ICK\emph{y} framework with corresponding kernel functions. It can be observed that ICK\emph{y} yields both smallest error and highest correlation with information from all 3 different sources. Hence, ICK\emph{y} works well with the $M = 3$ case and the regression performance is improved as we add in more information related to the target.}
\newpage

\subsubsection{Cases when \texorpdfstring{$M > 3$}{M > 3}}
\label{sec:5.1.3}

\begin{wraptable}{R}{0.53\textwidth}
\begin{minipage}{0.53\textwidth}
\vspace{-11mm}
\begin{table}[H]
\caption{Prediction errors of ICK\emph{y} on synthetic dataset generated by GP with multiplicative kernel, i.e., $y \sim \mathcal{GP} \left( 0, \prod_{i=1}^{M} K_{i} \right)$.}
\vspace{3mm}
\begin{tabular}{c|ccccc} 
\hline
     & $M=1$    & $M=2$    & $M=3$    & $M=4$    & $M=5$     \\ 
\hline
RMSE & 0.0334 & 0.0440 & 0.0466 & 0.2108 & 0.2560  \\
MAE  & 0.0266 & 0.0347 & 0.0353 & 0.1565 & 0.1790  \\
\hline
\end{tabular}
\label{tab:stress}
\vspace{-2mm}
\end{table}
\end{minipage}
\end{wraptable}

\textcolor{black}{To assess the robustness of ICK\emph{y} in handling diverse information sources, we conduct ``stress tests'' by evaluating its performance under varying numbers of information sources, i.e., when $M \in \{1,2,3,4,5\}$. In this analysis, we generate synthetic data following a methodology similar to that outlined in Section \ref{sec:5.1}. Specifically, we model the data as $y \sim \mathcal{GP} \left( 0, \prod_{i=1}^{M} K_{i} \right)$, where $K_1$, $K_2$, $K_3$, $K_4$, and $K_5$ are set to be the spectral mixture kernel, linear kernel, RBF kernel, exponential sine-squared kernel, and Mat\'{e}rn kernel with $\nu = 5/2$, respectively. As presented in Table \ref{tab:stress}, when $M > 3$, the performance of ICK\emph{y} starts to degrade. We posit that this degradation is primarily attributed to vanishing gradient, a phenomenon we will discuss further in the Discussion section.}

\subsection{Remote Sensing Data}
\label{sec:5.2}

We believe ICK\emph{y} will be particularly useful for remote sensing applications. \textcolor{black}{In this experiment, we evaluate ICK\emph{y} on a remote sensing datasets where each data point $\boldsymbol{x} = \{x, t\}$ contains 2 sources of information: a three-band natural color (red-blue-green) satellite image $x$ as the high-dimensional information and the corresponding timestamp as the low-dimensional information. Our goal is to forecast the ground-level PM$_{2.5}$ concentration $\hat{y} = f(x, t)$ using both sources of information.} 

As PM$_{2.5}$ varies with time on a yearly basis, we use an \emph{exponential-sine-squared} kernel with a period of $T = 365$ (days) to process the low-dimensional information $t$. The satellite images are processed with a CNN. The results of ICK\emph{y} are then compared with 2 benchmarks as shown in Table \ref{tab:1}: a Convolutional Neural Network-Random Forest (CNN-RF) joint model \citep{zheng2020estimating,zheng2021local} (similar to the MLP-RF model in Section \ref{sec:5.1}) and a \emph{carefully designed} Seasonal CNN-RF model that maps $t$ into two new features, $\sin(2\pi t/365)$ and $\cos(2\pi t/365)$, to explicitly model seasonality.

\begin{wraptable}{R}{0.53\textwidth}
\begin{minipage}{0.53\textwidth}
\vspace{-8mm}
\begin{table}[H]
\caption{Correlation and error statistics of ICK\emph{y} and other joint deep models with both convolutional and attention-based architectures on the PM$_{2.5}$ forecasting task. ``S.'' denotes seasonal variants.}
\vspace{3mm}
\begin{tabular}{c|ccc|c} 
\hline
~                                                             & R$_{\text{Spear}}$  & RMSE & MAE  & MSLL \\ 
\hline
CNN-RF                                                       & 0.00       & 194.63 & 185.83 & -     \\
ViT-RF                                                     & 0.07       & 190.82 & 181.63 & -     \\
\begin{tabular}[c]{@{}c@{}}S. CNN-RF\end{tabular}     & 0.62       & 53.36 & 39.38 & 96.77     \\
\begin{tabular}[c]{@{}c@{}}S. ViT-RF\end{tabular}     & 0.66       & 56.45 & 41.73 & 14.69     \\
\begin{tabular}[c]{@{}c@{}}S. Deep-ViT-RF 
%\cite{zhou2021deepvit}
\end{tabular} & 0.65       & 56.36 & 42.46 & 17.63     \\
\begin{tabular}[c]{@{}c@{}}S. MAE-ViT-RF 
%\cite{he2022masked}
\end{tabular} & 0.67       & 53.87 & 40.78 & 31.09     \\ 
\hline
CNN-ICK\emph{y}                                                      & 0.62       & 53.46 & 39.76 & \textbf{10.92}     \\
ViT-ICK\emph{y}                                                      & \textbf{0.68}       & 56.56 & 41.41 & 12208     \\
DeepViT-ICK\emph{y}                                                  & 0.66       & \textbf{52.41} & \textbf{35.93} & 38220     \\
\hline
\end{tabular}
\label{tab:1}
% \vspace{-4mm}
\end{table}
\end{minipage}
\end{wraptable}

We note that the inner product operation in ICK is similar in mathematical structure to attention-based mechanisms \citep{vaswani2017attention}, which are popular in many deep learning frameworks. Therefore, we introduce 2 variants of ICK\emph{y} and compare them with another 4 benchmarks where the CNN is replaced by an attention-based mechanism based off a Vision Transformer (ViT) \citep{dosovitskiy2020image} and DeepViT \citep{zhou2021deepvit} architectures.  These models include ViT-RF, Seasonal ViT-RF, Seasonal DeepViT-RF, and Seasonal MAE-ViT-RF, where ViT is pre-trained by a Masked Autoencoder \citep{he2022masked}.  ``Seasonal'' here denotes that we use the transformed periodic representation defined previously. We note that we are unaware of Vision Transformers being used in this manner, and we want to evaluate whether it is the model structure or the prior information that is improving the results. Hence, these attention-based models, while not a primary contribution, are novel and represent a good-faith effort to define models with similar forms. As displayed in Table \ref{tab:1}, both standard CNN-RF and ViT-RF models yields very large errors on predicting PM$_{2.5}$ values. After including seasonality, CNN-RF performs significantly better and shares similar predictive performance with CNN-ICK\emph{y}. After replacing CNN with ViT, we observe slight improvement in both RF-joint models (especially when pre-trained by MAE) and ICK\emph{y} variants. Among the models we present here, ViT-ICK\emph{y} and DeepViT-ICK\emph{y} achieve the highest correlation and the smallest prediction error, respectively. To evaluate the uncertainty calibration capability of ICK\emph{y}, we construct \textbf{ICK\emph{y} ensembles} by following Algorithm \ref{alg:2} and compare them with ensemble formulations of RF-joint model benchmarks. We use a criterion called Mean Standardized Log Loss (MSLL) as defined in \citet{williams2006gaussian} to evaluate the uncertainty calibration. From Table \ref{tab:1}, it can be observed that CNN-ICK\emph{y} ensemble achieves the smallest MSLL. In addition, we realize ViT-based ICK\emph{y} variants yield very large MSLL as they make predictions which are far from the true PM$_{2.5}$ labels with high confidence (i.e. small variance). \textcolor{black}{We provide details and possible explanations in Appendix \ref{appx:H}.}

\subsection{Other Regression Datasets}
\label{sec:5.3}

\textcolor{black}{To see if our ICK\emph{y} framework generalizes to other domains, we acquired two additional regression datasets from the UCI machine learning repository, one for predicting worker productivity (with 1197 samples and 15 features) and the other for predicting power consumption (with 2075259 samples and 9 features). For the worker productivity dataset, we employ ICK\emph{y} with an \emph{exponential-sine-squared} kernel with different periods and compare them with benchmarks including MLPs \citep{al2019deep} and Deep Kernel Learning (DKL) \citep{wilson2016deep} as shown in Table \ref{tab:2}. For the power consumption dataset, we use ICK\emph{y} with both an \emph{exponential-sine-squared} kernel and a \emph{Mat\'ern 3/2} kernel and compare them with scalable exact GP \citep{wang2019exact} and stochastic variational GP \citep{hensman2013gaussian} as shown in Table \ref{tab:3}. From the results, we observe that ICK\emph{y} outperforms all the benchmarks in both experiments (especially on the worker productivity data with a margin of almost one order of magnitude).}

\begin{table}[t!]
\begin{minipage}[c]{0.48\textwidth}
% \begin{table}[H]
\centering
\caption{Prediction error of actual worker productivity on the test data set with ICK\emph{y} and other benchmark models.}
\vspace{3mm}
\resizebox{\textwidth}{!}{
\begin{tabular}{c|cc} 
\hline
~            & MSE $\downarrow (*10^{-3})$ & MAE $\downarrow (*10^{-2})$ \\ 
\hline
MLP         & 20.16 $\pm$ 1.26   & 9.93 $\pm$ 0.36  \\
Cyclic MLP   & 20.97 $\pm$ 1.98   & 10.16 $\pm$ 0.77   \\
Cyclic MLP-RF   & 19.05 $\pm$ 1.36   & 9.70 $\pm$ 0.48   \\
DKL          & 21.40 $\pm$ 2.85   & 11.14 $\pm$ 0.94      \\
\hline
ICK\emph{y}, $T = 2$  & 3.43 $\pm$ 1.42   & 4.85 $\pm$ 1.00   \\
ICK\emph{y}, $T = 7$  & 0.44 $\pm$ 0.13   & 1.43 $\pm$ 0.15   \\
ICK\emph{y}, $T = 30$ & \textbf{0.31} $\pm$ 0.09   & \textbf{1.17} $\pm$ 0.14  \\
\hline
\end{tabular}}
\label{tab:2}
\vspace{-2mm}
% \end{table}
\end{minipage}\hfill
\begin{minipage}[c]{0.48\textwidth}
% \begin{table}[H]
\centering
\caption{Root-mean-square-error (RMSE) and negative log-likelihood (NLL) of ICK\emph{y} with Matern-3/2 kernel and periodic kernel and two GP benchmarks on power consumption data.}
\vspace{3mm}
\resizebox{\textwidth}{!}{
\begin{tabular}{c|cc} 
\hline
                 & RMSE                               & NLL                                  \\ 
\hline
Exact GP         & 0.055 $\pm$ 0.000 & -0.152 $\pm$ 0.001  \\
SVGP             & 0.084 $\pm$ 0.005 & -1.010 $\pm$ 0.039  \\ 
\hline
ICK\emph{y}, Mat\'ern-3/2 & 0.036 $\pm$ 0.000 & -1.424 $\pm$ 0.000  \\
ICK\emph{y}, periodic   & \textbf{0.033} $\pm$ 0.000 & \textbf{-1.737} $\pm$ 0.000  \\
\hline
\end{tabular}}
\label{tab:3}
% \end{table}
\end{minipage}
\end{table}

\subsection{Adapting ICK for Classification}
\label{sec:5.4}

While regression tasks are the primary motivation for this paper, there are many ways to adapt GPR for classification tasks.  For example, a binary classification model can be created by using a sigmoid \citep{williams1998bayesian} or probit link \citep{choudhuri2007nonparametric} on the output of the GP.  Succinctly, given a function $f(\boldsymbol{x}) \sim \mathcal{GP}\left(0, K(\boldsymbol{x}, \boldsymbol{x}')\right)$, the binary outcome probability is be given as $p(y=1|f(x))=\sigma(f(x))$.  Likewise, a multiple classification model can be constructed by using a multi-output GP (or multiple GPs) and putting the outputs through a softmax function \citep{williams1998bayesian} or multinomial probit link \citep{girolami2006variational}.  This strategy can be summarized by calculating $C$ different functions $f_c(\boldsymbol{x})\sim \mathcal{GP}\left(0, K(\boldsymbol{x}, \boldsymbol{x}')\right)$ for $c=1,..., C$, where $C$ is the number of classes, and then calculating the class probabilities through a link function,  $p(y|\boldsymbol{x})=\text{softmax}([f_1(\boldsymbol{x}),f_2(\boldsymbol{x}),...,f_C(\boldsymbol{x})])$.

This same logic can be used to construct a multiple classification model from ICK\emph{y}.  Succinctly, let $r_c=f_{\text{NN},c}(\boldsymbol{x}^{(1)})\odot \boldsymbol{z}^{(2)}_c$, where $f_{\text{NN},c}$ denotes a neural network specific to the $c^{th}$ class and $z^{(2)}_c$ represents the Nyström approximation specific to the kernel for the $c^{th}$ class. We note that often in a multi-output case the kernel parameters are shared, and so $\boldsymbol{z}^{(2)}_c$ would be an identical vector for each class.  Then, the output probabilities for a data sample as $p(y|\boldsymbol{x})=\text{softmax}([r_1,\dots,r_C])$.  This framework is learned with a cross-entropy loss.

To provide \textit{proof-of-concept} of this multiple classification strategy, we implemented this model on a version of Rotating MNIST.  In this task, a dataset was created by rotating each image in the dataset by a uniform random value $\phi\in[0,2\pi)$, thus creating a dataset with 60,000 images each with an associated rotation covariate $\phi$.  We implemented the above multiple classification model with a periodic kernel over the rotation angle.  This strategy yielded an accuracy of 92.3\% on the validation data.  This is lower than methods such as spatial transformers \citep{jaderberg2015spatial} that report accuracy greater than 99\%.  However, those models explicitly use the fact that the information is simply rotated, whereas ICK is modeling a smooth transformation in the prediction function as a function of angle.   This ICK classification model is much closer in concept to the way Rotating MNIST is used to evaluate unsupervised domain adaptation.  While the evaluation strategy is different than our random validation set, the state-of-the-art accuracy on unsupervised domain adaption is 87.1\% \citep{wang2020continuously}.  Due to the lack of complete and fair comparisons, we are not claiming that ICK\emph{y} is state-of-the-art for classification, but ICK\emph{y}'s classification model does seem reasonable and viable based upon this result.

\section{Discussion}
\label{sec:6}

\paragraph{Computational Complexity and Flexibility of ICK}
Compared to exact composite GP models which scale $\mathcal{O}(N^3)$, the training process of our ICK framework is more efficient as it leverages standard backpropagation to learn both the paramters of NN and the kernel function. \textcolor{black}{Specifically, let $N_B$
be the number of data points in one mini-batch and $p$ be the number of inducing points for Nystr\"om approximation. The computational complexity for $g(\cdot)$ is $\mathcal{O}(p^2 N_B + p^3)$. The computational complexity for the forward pass of $f_{\text{NN}}(\cdot)$ is $\mathcal{O}\left( N_B(k_1 k_2 + \cdots k_{L-1} k_L) \right)$ where $k_1, ..., k_L$ are the widths of each layer for a neural network of depth $L$.} In addition, our ICK framework is more flexible compared to other joint models (i.e. BNNs and CNN-RF). Specifically, the BNNs of \citet{pearce2020expressive} cannot simulate complicated kernels such as the spectral mixture kernel we use in Section \ref{sec:5.1}. 

\paragraph{Limitations}
% A limitation of ICK lies in our method of combining latent representations as the nature of inner product (i.e. the effect of multiplying small numbers) may cause \emph{vanishing gradient} problems when we have a large number of sources of information (i.e. $M$ is large). 
As described in Section \ref{sec:5.1.3}, ICK\emph{y} starts to yield higher prediction error when we have more than 3 different sources of information. This phenomenon is likely to be caused by vanishing gradient as we may need to multiply small numbers together due to the nature of inner product.
Furthermore, in some cases, the properties of predictive posterior given by ICK ensemble are dominated by the neural network as shown in Appendix \ref{appx:H}. It can sometimes be challenging to choose an appropriate NN architecture to make sure it does not interfere with other specified kernels. \textcolor{black}{In addition, as highlighted in Section 8.3.2 of \citet{williams2006gaussian}, it is important to note that the predictive variance using the Nystr\"{o}m method is not guaranteed to be positive. This can result in the kernel matrix no longer being positive-semidefinite, making the Cholesky decomposition infeasible. In such cases, we recommend either reducing the rank of the Nystr\"{o}m approximation or experimenting with alternative kernel functions.}

\paragraph{Broader Impacts}
We believe our framework is extensively applicable to regression problems in many fields of study involving high-dimensional data and multiple sources of information with perceptible trends, such as remote sensing, spatial statistics, or clinical diagnosis. Also, we are not aware of any negative societal impacts of our work.

\section{Conclusion}
\label{sec:7}
This paper presents a novel yet surprisingly simple Implicit Composite Kernel (ICK) framework to learn from \emph{hybrid} data containing both high-dimensional information and low-dimensional information with prior knowledge. We first analytically show the resemblance between ICK and composite GPR models and then conduct experiments using both synthetic and real-world data. It appears that ICK outperforms various benchmark models in our experiments with lowest prediction errors and highest correlations even with very limited data. Overall, we show that our ICK framework is exceptionally powerful when learning from \emph{hybrid} data with prior knowledge incorporated, and we hope our work can inspire more future research on joint machine learning models, enhancing their performance, efficiency, flexibility, and generalization capability.

\bibliography{main}

\begin{thebibliography}{63}
\providecommand{\natexlab}[1]{#1}
\providecommand{\url}[1]{\texttt{#1}}
\expandafter\ifx\csname urlstyle\endcsname\relax
  \providecommand{\doi}[1]{doi: #1}\else
  \providecommand{\doi}{doi: \begingroup \urlstyle{rm}\Url}\fi

\bibitem[Adlam et~al.(2020)Adlam, Lee, Xiao, Pennington, and Snoek]{adlam2020exploring}
Ben Adlam, Jaehoon Lee, Lechao Xiao, Jeffrey Pennington, and Jasper Snoek.
\newblock Exploring the uncertainty properties of neural networks' implicit priors in the infinite-width limit.
\newblock \emph{arXiv preprint arXiv:2010.07355}, 2020.

\bibitem[Adlam et~al.(2023)Adlam, Lee, Padhy, Nado, and Snoek]{adlam2023kernel}
Ben Adlam, Jaehoon Lee, Shreyas Padhy, Zachary Nado, and Jasper Snoek.
\newblock Kernel regression with infinite-width neural networks on millions of examples.
\newblock \emph{arXiv preprint arXiv:2303.05420}, 2023.

\bibitem[Al~Imran et~al.(2019)Al~Imran, Amin, Rifat, and Mehreen]{al2019deep}
Abdullah Al~Imran, Md~Nur Amin, Md~Rifatul~Islam Rifat, and Shamprikta Mehreen.
\newblock Deep neural network approach for predicting the productivity of garment employees.
\newblock In \emph{2019 6th International Conference on Control, Decision and Information Technologies (CoDIT)}, pp.\  1402--1407. IEEE, 2019.

\bibitem[Bishop \& Nasrabadi(2006)Bishop and Nasrabadi]{bishop2006pattern}
Christopher~M Bishop and Nasser~M Nasrabadi.
\newblock \emph{Pattern recognition and machine learning}, volume~4.
\newblock Springer, 2006.

\bibitem[Bottou et~al.(2018)Bottou, Curtis, and Nocedal]{bottou2018optimization}
L{\'e}on Bottou, Frank~E Curtis, and Jorge Nocedal.
\newblock Optimization methods for large-scale machine learning.
\newblock \emph{Siam Review}, 60\penalty0 (2):\penalty0 223--311, 2018.

\bibitem[Bouhlel \& Martins(2019)Bouhlel and Martins]{bouhlel2019gradient}
Mohamed~A Bouhlel and Joaquim~RRA Martins.
\newblock Gradient-enhanced kriging for high-dimensional problems.
\newblock \emph{Engineering with Computers}, 35\penalty0 (1):\penalty0 157--173, 2019.

\bibitem[Bouhlel et~al.(2016)Bouhlel, Bartoli, Otsmane, and Morlier]{bouhlel2016improving}
Mohamed~Amine Bouhlel, Nathalie Bartoli, Abdelkader Otsmane, and Joseph Morlier.
\newblock Improving kriging surrogates of high-dimensional design models by partial least squares dimension reduction.
\newblock \emph{Structural and Multidisciplinary Optimization}, 53\penalty0 (5):\penalty0 935--952, 2016.

\bibitem[Choudhuri et~al.(2007)Choudhuri, Ghosal, and Roy]{choudhuri2007nonparametric}
Nidhan Choudhuri, Subhashis Ghosal, and Anindya Roy.
\newblock Nonparametric binary regression using a gaussian process prior.
\newblock \emph{Statistical Methodology}, 4\penalty0 (2):\penalty0 227--243, 2007.

\bibitem[Datta et~al.(2016)Datta, Banerjee, Finley, and Gelfand]{datta2016hierarchical}
Abhirup Datta, Sudipto Banerjee, Andrew~O Finley, and Alan~E Gelfand.
\newblock Hierarchical nearest-neighbor gaussian process models for large geostatistical datasets.
\newblock \emph{Journal of the American Statistical Association}, 111\penalty0 (514):\penalty0 800--812, 2016.

\bibitem[Dosovitskiy et~al.(2020)Dosovitskiy, Beyer, Kolesnikov, Weissenborn, Zhai, Unterthiner, Dehghani, Minderer, Heigold, Gelly, et~al.]{dosovitskiy2020image}
Alexey Dosovitskiy, Lucas Beyer, Alexander Kolesnikov, Dirk Weissenborn, Xiaohua Zhai, Thomas Unterthiner, Mostafa Dehghani, Matthias Minderer, Georg Heigold, Sylvain Gelly, et~al.
\newblock An image is worth 16x16 words: Transformers for image recognition at scale.
\newblock \emph{arXiv preprint arXiv:2010.11929}, 2020.

\bibitem[Drineas et~al.(2005)Drineas, Mahoney, and Cristianini]{drineas2005nystrom}
Petros Drineas, Michael~W Mahoney, and Nello Cristianini.
\newblock On the nystr{\"o}m method for approximating a gram matrix for improved kernel-based learning.
\newblock \emph{journal of machine learning research}, 6\penalty0 (12), 2005.

\bibitem[Duvenaud(2014)]{duvenaud2014automatic}
David Duvenaud.
\newblock \emph{Automatic model construction with Gaussian processes}.
\newblock PhD thesis, University of Cambridge, 2014.

\bibitem[Garnelo et~al.(2018)Garnelo, Schwarz, Rosenbaum, Viola, Rezende, Eslami, and Teh]{garnelo2018neural}
Marta Garnelo, Jonathan Schwarz, Dan Rosenbaum, Fabio Viola, Danilo~J Rezende, SM~Eslami, and Yee~Whye Teh.
\newblock Neural processes.
\newblock \emph{arXiv preprint arXiv:1807.01622}, 2018.

\bibitem[Garriga-Alonso et~al.(2018)Garriga-Alonso, Rasmussen, and Aitchison]{garriga2018deep}
Adri{\`a} Garriga-Alonso, Carl~Edward Rasmussen, and Laurence Aitchison.
\newblock Deep convolutional networks as shallow gaussian processes.
\newblock \emph{arXiv preprint arXiv:1808.05587}, 2018.

\bibitem[Gelfand \& Schliep(2016)Gelfand and Schliep]{gelfand2016spatial}
Alan~E Gelfand and Erin~M Schliep.
\newblock Spatial statistics and gaussian processes: A beautiful marriage.
\newblock \emph{Spatial Statistics}, 18:\penalty0 86--104, 2016.

\bibitem[Girolami \& Rogers(2006)Girolami and Rogers]{girolami2006variational}
Mark Girolami and Simon Rogers.
\newblock Variational bayesian multinomial probit regression with gaussian process priors.
\newblock \emph{Neural Computation}, 18\penalty0 (8):\penalty0 1790--1817, 2006.

\bibitem[Goodfellow et~al.(2016)Goodfellow, Bengio, and Courville]{goodfellow2016deep}
Ian Goodfellow, Yoshua Bengio, and Aaron Courville.
\newblock \emph{Deep learning}.
\newblock MIT press, 2016.

\bibitem[He et~al.(2020)He, Lakshminarayanan, and Teh]{he2020bayesian}
Bobby He, Balaji Lakshminarayanan, and Yee~Whye Teh.
\newblock Bayesian deep ensembles via the neural tangent kernel.
\newblock \emph{Advances in neural information processing systems}, 33:\penalty0 1010--1022, 2020.

\bibitem[He et~al.(2022)He, Chen, Xie, Li, Doll{\'a}r, and Girshick]{he2022masked}
Kaiming He, Xinlei Chen, Saining Xie, Yanghao Li, Piotr Doll{\'a}r, and Ross Girshick.
\newblock Masked autoencoders are scalable vision learners.
\newblock In \emph{Proceedings of the IEEE/CVF Conference on Computer Vision and Pattern Recognition}, pp.\  16000--16009, 2022.

\bibitem[Hensman et~al.(2013)Hensman, Fusi, and Lawrence]{hensman2013gaussian}
James Hensman, Nicolo Fusi, and Neil~D Lawrence.
\newblock Gaussian processes for big data.
\newblock \emph{arXiv preprint arXiv:1309.6835}, 2013.

\bibitem[Hinton \& Salakhutdinov(2007)Hinton and Salakhutdinov]{hinton2007using}
Geoffrey~E Hinton and Russ~R Salakhutdinov.
\newblock Using deep belief nets to learn covariance kernels for gaussian processes.
\newblock \emph{Advances in neural information processing systems}, 20, 2007.

\bibitem[Jacot et~al.(2018)Jacot, Gabriel, and Hongler]{jacot2018neural}
Arthur Jacot, Franck Gabriel, and Cl{\'e}ment Hongler.
\newblock Neural tangent kernel: Convergence and generalization in neural networks.
\newblock \emph{Advances in neural information processing systems}, 31, 2018.

\bibitem[Jaderberg et~al.(2015)Jaderberg, Simonyan, Zisserman, et~al.]{jaderberg2015spatial}
Max Jaderberg, Karen Simonyan, Andrew Zisserman, et~al.
\newblock Spatial transformer networks.
\newblock \emph{Advances in neural information processing systems}, 28, 2015.

\bibitem[Jiang et~al.(2022)Jiang, Zheng, Bergin, and Carlson]{jiang2022improving}
Ziyang Jiang, Tongshu Zheng, Mike Bergin, and David Carlson.
\newblock Improving spatial variation of ground-level pm2. 5 prediction with contrastive learning from satellite imagery.
\newblock \emph{Science of Remote Sensing}, pp.\  100052, 2022.

\bibitem[Journel \& Huijbregts(1976)Journel and Huijbregts]{journel1976mining}
Andre~G Journel and Charles~J Huijbregts.
\newblock \emph{Mining geostatistics}.
\newblock The Blackburn Press, 1976.

\bibitem[Kim et~al.(2005)Kim, Mallick, and Holmes]{kim2005analyzing}
Hyoung-Moon Kim, Bani~K Mallick, and Chris~C Holmes.
\newblock Analyzing nonstationary spatial data using piecewise gaussian processes.
\newblock \emph{Journal of the American Statistical Association}, 100\penalty0 (470):\penalty0 653--668, 2005.

\bibitem[Krige(1951)]{krige1951statistical}
Daniel~G Krige.
\newblock A statistical approach to some basic mine valuation problems on the witwatersrand.
\newblock \emph{Journal of the Southern African Institute of Mining and Metallurgy}, 52\penalty0 (6):\penalty0 119--139, 1951.

\bibitem[Lagaris et~al.(1998)Lagaris, Likas, and Fotiadis]{lagaris1998artificial}
Isaac~E Lagaris, Aristidis Likas, and Dimitrios~I Fotiadis.
\newblock Artificial neural networks for solving ordinary and partial differential equations.
\newblock \emph{IEEE transactions on neural networks}, 9\penalty0 (5):\penalty0 987--1000, 1998.

\bibitem[LeCun et~al.(2015)LeCun, Bengio, and Hinton]{lecun2015deep}
Yann LeCun, Yoshua Bengio, and Geoffrey Hinton.
\newblock Deep learning.
\newblock \emph{nature}, 521\penalty0 (7553):\penalty0 436--444, 2015.

\bibitem[Lee et~al.(2017)Lee, Bahri, Novak, Schoenholz, Pennington, and Sohl-Dickstein]{lee2017deep}
Jaehoon Lee, Yasaman Bahri, Roman Novak, Samuel~S Schoenholz, Jeffrey Pennington, and Jascha Sohl-Dickstein.
\newblock Deep neural networks as gaussian processes.
\newblock \emph{arXiv preprint arXiv:1711.00165}, 2017.

\bibitem[Lee et~al.(2019)Lee, Xiao, Schoenholz, Bahri, Novak, Sohl-Dickstein, and Pennington]{lee2019wide}
Jaehoon Lee, Lechao Xiao, Samuel Schoenholz, Yasaman Bahri, Roman Novak, Jascha Sohl-Dickstein, and Jeffrey Pennington.
\newblock Wide neural networks of any depth evolve as linear models under gradient descent.
\newblock \emph{Advances in neural information processing systems}, 32, 2019.

\bibitem[Maddison et~al.(2016)Maddison, Mnih, and Teh]{maddison2016concrete}
Chris~J Maddison, Andriy Mnih, and Yee~Whye Teh.
\newblock The concrete distribution: A continuous relaxation of discrete random variables.
\newblock \emph{arXiv preprint arXiv:1611.00712}, 2016.

\bibitem[Marcus(2018)]{marcus2018deep}
Gary Marcus.
\newblock Deep learning: A critical appraisal.
\newblock \emph{arXiv preprint arXiv:1801.00631}, 2018.

\bibitem[Matsubara et~al.(2020)Matsubara, Oates, and Briol]{matsubara2020ridgelet}
Takuo Matsubara, Chris~J Oates, and Fran{\c{c}}ois-Xavier Briol.
\newblock The ridgelet prior: A covariance function approach to prior specification for bayesian neural networks.
\newblock \emph{arXiv preprint arXiv:2010.08488}, 2020.

\bibitem[Matthews et~al.(2017)Matthews, Hron, Turner, and Ghahramani]{matthews2017sample}
Alexander G de~G Matthews, Jiri Hron, Richard~E Turner, and Zoubin Ghahramani.
\newblock Sample-then-optimize posterior sampling for bayesian linear models.
\newblock In \emph{NeurIPS Workshop on Advances in Approximate Bayesian Inference}, 2017.

\bibitem[Matthews et~al.(2018)Matthews, Rowland, Hron, Turner, and Ghahramani]{matthews2018gaussian}
Alexander G de~G Matthews, Mark Rowland, Jiri Hron, Richard~E Turner, and Zoubin Ghahramani.
\newblock Gaussian process behaviour in wide deep neural networks.
\newblock \emph{arXiv preprint arXiv:1804.11271}, 2018.

\bibitem[McFee et~al.(2011)McFee, Lanckriet, and Jebara]{mcfee2011learning}
Brian McFee, Gert Lanckriet, and Tony Jebara.
\newblock Learning multi-modal similarity.
\newblock \emph{Journal of machine learning research}, 12\penalty0 (2), 2011.

\bibitem[Moseley et~al.(2020)Moseley, Markham, and Nissen-Meyer]{moseley2020solving}
Ben Moseley, Andrew Markham, and Tarje Nissen-Meyer.
\newblock Solving the wave equation with physics-informed deep learning.
\newblock \emph{arXiv preprint arXiv:2006.11894}, 2020.

\bibitem[Neal(1996)]{neal1996priors}
Radford~M Neal.
\newblock Priors for infinite networks.
\newblock In \emph{Bayesian Learning for Neural Networks}, pp.\  29--53. Springer, 1996.

\bibitem[Novak et~al.(2018)Novak, Xiao, Lee, Bahri, Yang, Hron, Abolafia, Pennington, and Sohl-Dickstein]{novak2018bayesian}
Roman Novak, Lechao Xiao, Jaehoon Lee, Yasaman Bahri, Greg Yang, Jiri Hron, Daniel~A Abolafia, Jeffrey Pennington, and Jascha Sohl-Dickstein.
\newblock Bayesian deep convolutional networks with many channels are gaussian processes.
\newblock \emph{arXiv preprint arXiv:1810.05148}, 2018.

\bibitem[Novak et~al.(2020)Novak, Xiao, Hron, Lee, Alemi, Sohl-Dickstein, and Schoenholz]{neuraltangents2020}
Roman Novak, Lechao Xiao, Jiri Hron, Jaehoon Lee, Alexander~A. Alemi, Jascha Sohl-Dickstein, and Samuel~S. Schoenholz.
\newblock Neural tangents: Fast and easy infinite neural networks in python.
\newblock In \emph{International Conference on Learning Representations}, 2020.
\newblock URL \url{https://github.com/google/neural-tangents}.

\bibitem[Pearce et~al.(2020)Pearce, Tsuchida, Zaki, Brintrup, and Neely]{pearce2020expressive}
Tim Pearce, Russell Tsuchida, Mohamed Zaki, Alexandra Brintrup, and Andy Neely.
\newblock Expressive priors in bayesian neural networks: Kernel combinations and periodic functions.
\newblock In \emph{Uncertainty in artificial intelligence}, pp.\  134--144. PMLR, 2020.

\bibitem[Petelin et~al.(2013)Petelin, Grancharova, and Kocijan]{petelin2013evolving}
Dejan Petelin, Alexandra Grancharova, and Ju{\v{s}} Kocijan.
\newblock Evolving gaussian process models for prediction of ozone concentration in the air.
\newblock \emph{Simulation modelling practice and theory}, 33:\penalty0 68--80, 2013.

\bibitem[Rahimi \& Recht(2007)Rahimi and Recht]{rahimi2007random}
Ali Rahimi and Benjamin Recht.
\newblock Random features for large-scale kernel machines.
\newblock \emph{Advances in neural information processing systems}, 20, 2007.

\bibitem[Rahimi \& Recht(2008)Rahimi and Recht]{rahimi2008weighted}
Ali Rahimi and Benjamin Recht.
\newblock Weighted sums of random kitchen sinks: Replacing minimization with randomization in learning.
\newblock \emph{Advances in neural information processing systems}, 21, 2008.

\bibitem[Snelson \& Ghahramani(2005)Snelson and Ghahramani]{snelson2005sparse}
Edward Snelson and Zoubin Ghahramani.
\newblock Sparse gaussian processes using pseudo-inputs.
\newblock \emph{Advances in neural information processing systems}, 18, 2005.

\bibitem[Titsias(2009)]{titsias2009variational}
Michalis Titsias.
\newblock Variational learning of inducing variables in sparse gaussian processes.
\newblock In \emph{Artificial intelligence and statistics}, pp.\  567--574. PMLR, 2009.

\bibitem[Van~der Wilk et~al.(2017)Van~der Wilk, Rasmussen, and Hensman]{van2017convolutional}
Mark Van~der Wilk, Carl~Edward Rasmussen, and James Hensman.
\newblock Convolutional gaussian processes.
\newblock \emph{Advances in Neural Information Processing Systems}, 30, 2017.

\bibitem[Vaswani et~al.(2017)Vaswani, Shazeer, Parmar, Uszkoreit, Jones, Gomez, Kaiser, and Polosukhin]{vaswani2017attention}
Ashish Vaswani, Noam Shazeer, Niki Parmar, Jakob Uszkoreit, Llion Jones, Aidan~N Gomez, {\L}ukasz Kaiser, and Illia Polosukhin.
\newblock Attention is all you need.
\newblock \emph{Advances in neural information processing systems}, 30, 2017.

\bibitem[Wang et~al.(2020)Wang, He, and Katabi]{wang2020continuously}
Hao Wang, Hao He, and Dina Katabi.
\newblock Continuously indexed domain adaptation.
\newblock \emph{International Conference on Machine Learning}, 2020.

\bibitem[Wang et~al.(2019)Wang, Pleiss, Gardner, Tyree, Weinberger, and Wilson]{wang2019exact}
Ke~Wang, Geoff Pleiss, Jacob Gardner, Stephen Tyree, Kilian~Q Weinberger, and Andrew~Gordon Wilson.
\newblock Exact gaussian processes on a million data points.
\newblock \emph{Advances in neural information processing systems}, 32, 2019.

\bibitem[Williams \& Seeger(2000)Williams and Seeger]{williams2000using}
Christopher Williams and Matthias Seeger.
\newblock Using the nystr{\"o}m method to speed up kernel machines.
\newblock \emph{Advances in neural information processing systems}, 13, 2000.

\bibitem[Williams \& Rasmussen(2006)Williams and Rasmussen]{williams2006gaussian}
Christopher~K Williams and Carl~Edward Rasmussen.
\newblock \emph{Gaussian processes for machine learning}, volume~2.
\newblock MIT press Cambridge, MA, 2006.

\bibitem[Williams \& Barber(1998)Williams and Barber]{williams1998bayesian}
Christopher~KI Williams and David Barber.
\newblock Bayesian classification with gaussian processes.
\newblock \emph{IEEE Transactions on pattern analysis and machine intelligence}, 20\penalty0 (12):\penalty0 1342--1351, 1998.

\bibitem[Wilson \& Adams(2013)Wilson and Adams]{wilson2013gaussian}
Andrew Wilson and Ryan Adams.
\newblock Gaussian process kernels for pattern discovery and extrapolation.
\newblock In \emph{International conference on machine learning}, pp.\  1067--1075. PMLR, 2013.

\bibitem[Wilson et~al.(2016{\natexlab{a}})Wilson, Hu, Salakhutdinov, and Xing]{wilson2016stochastic}
Andrew~G Wilson, Zhiting Hu, Russ~R Salakhutdinov, and Eric~P Xing.
\newblock Stochastic variational deep kernel learning.
\newblock \emph{Advances in neural information processing systems}, 29, 2016{\natexlab{a}}.

\bibitem[Wilson et~al.(2011)Wilson, Knowles, and Ghahramani]{wilson2011gaussian}
Andrew~Gordon Wilson, David~A Knowles, and Zoubin Ghahramani.
\newblock Gaussian process regression networks.
\newblock \emph{arXiv preprint arXiv:1110.4411}, 2011.

\bibitem[Wilson et~al.(2016{\natexlab{b}})Wilson, Hu, Salakhutdinov, and Xing]{wilson2016deep}
Andrew~Gordon Wilson, Zhiting Hu, Ruslan Salakhutdinov, and Eric~P Xing.
\newblock Deep kernel learning.
\newblock In \emph{Artificial intelligence and statistics}, pp.\  370--378. PMLR, 2016{\natexlab{b}}.

\bibitem[Yang et~al.(2012)Yang, Li, Mahdavi, Jin, and Zhou]{yang2012nystrom}
Tianbao Yang, Yu-Feng Li, Mehrdad Mahdavi, Rong Jin, and Zhi-Hua Zhou.
\newblock Nystr{\"o}m method vs random fourier features: A theoretical and empirical comparison.
\newblock \emph{Advances in neural information processing systems}, 25, 2012.

\bibitem[Zhang et~al.(2011)Zhang, Wang, Zhou, Yuan, Shen, Initiative, et~al.]{zhang2011multimodal}
Daoqiang Zhang, Yaping Wang, Luping Zhou, Hong Yuan, Dinggang Shen, Alzheimer's Disease~Neuroimaging Initiative, et~al.
\newblock Multimodal classification of alzheimer's disease and mild cognitive impairment.
\newblock \emph{Neuroimage}, 55\penalty0 (3):\penalty0 856--867, 2011.

\bibitem[Zheng et~al.(2020)Zheng, Bergin, Hu, Miller, and Carlson]{zheng2020estimating}
Tongshu Zheng, Michael~H Bergin, Shijia Hu, Joshua Miller, and David~E Carlson.
\newblock Estimating ground-level pm2. 5 using micro-satellite images by a convolutional neural network and random forest approach.
\newblock \emph{Atmospheric Environment}, 230:\penalty0 117451, 2020.

\bibitem[Zheng et~al.(2021)Zheng, Bergin, Wang, and Carlson]{zheng2021local}
Tongshu Zheng, Michael Bergin, Guoyin Wang, and David Carlson.
\newblock Local pm2. 5 hotspot detector at 300 m resolution: A random forest--convolutional neural network joint model jointly trained on satellite images and meteorology.
\newblock \emph{Remote Sensing}, 13\penalty0 (7):\penalty0 1356, 2021.

\bibitem[Zhou et~al.(2021)Zhou, Kang, Jin, Yang, Lian, Jiang, Hou, and Feng]{zhou2021deepvit}
Daquan Zhou, Bingyi Kang, Xiaojie Jin, Linjie Yang, Xiaochen Lian, Zihang Jiang, Qibin Hou, and Jiashi Feng.
\newblock Deepvit: Towards deeper vision transformer.
\newblock \emph{arXiv preprint arXiv:2103.11886}, 2021.

\end{thebibliography}
\bibliographystyle{tmlr}

\newpage

\appendix
\section{Proof of Lemma 2}
\renewcommand\thefigure{\thesection\arabic{figure}}
\setcounter{figure}{0}
\label{appx:A}

\textcolor{black}{As stated in Section \ref{sec:4.1} and the previous section, Theorem \ref{thm:1} relies on Lemma \ref{lemma:2} which states that all entries in $\boldsymbol{z}^{(1)}$ (the representation extracted by the neural network) are independent with each other. To prove this lemma, we can explicitly write the $k^{th}$ entry in $\boldsymbol{z}^{(1)}$ as follows:}
\begin{equation}
z_k^{(1)}(\cdot) = b_k^{(1)} + \sum_{r=1}^{N_{z^{(1)}}} W_{kr}^{(1)} \phi\left({z'}^{(1)}_r(\cdot)\right),
\end{equation}
\textcolor{black}{where $W_{kr}^{(1)}$ and $b_k^{(1)}$ are i.i.d. Gaussian weight and bias parameters and ${z'}^{(1)}_r$ is the $r^{th}$ entry in the previous layer. We see that $z_k^{(1)}$ is a linear combination of i.i.d. Gaussian terms, and we can show that $z_k^{(1)}(x_i)$ and $z_l^{(1)}(x_j)$ are independent for two different inputs $x_i$ and $x_j$ and $k \neq l$ by computing the covariance as shown below.}
\begin{align*}
&\text{cov}\left( z_k^{(1)}(x_i), z_l^{(1)}(x_j) \right) \\
&= \textstyle \mathbb{E} \left[ \left( b_k^{(1)} + \sum_{r=1}^{N_{z^{(1)}}} W_{kr}^{(1)} \phi\left({z'}^{(1)}_r(x_i)\right) \right)
\left( b_l^{(1)} + \sum_{s=1}^{N_{z^{(1)}}} W_{ls}^{(1)} \phi\left({z'}^{(1)}_s(x_j)\right) \right) \right] \\
&= \mathbb{E}\left[ b_k^{(1)} b_l^{(1)} \right] + \mathbb{E}\left[ b_k^{(1)}\sum_{s=1}^{N_{z^{(1)}}} W_{ls}^{(1)} \phi\left({z'}^{(1)}_s(x_j)\right) \right] + \mathbb{E}\left[ b_l^{(1)}\sum_{r=1}^{N_{z^{(1)}}} W_{kr}^{(1)} \phi\left({z'}^{(1)}_r(x_i)\right) \right] \\
&+ \mathbb{E}\left[ \left( \sum_{r=1}^{N_{z^{(1)}}} W_{kr}^{(1)} \phi\left({z'}^{(1)}_r(x_i)\right) \right)
\left( \sum_{s=1}^{N_{z^{(1)}}} W_{ls}^{(1)} \phi\left({z'}^{(1)}_s(x_j)\right) \right) \right]
\end{align*}
Since the weights and biases are all i.i.d. Gaussian with zero mean \emph{a priori} and they are independent of the outputs from the previous layer, we have:
\begin{align*}
&\mathbb{E}\left[ b_k^{(1)} b_l^{(1)} \right] = \mathbb{E}\left[b_k^{(1)}\right] \mathbb{E}\left[b_l^{(1)}\right] = 0 \\
& \textstyle \mathbb{E}\left[ b_k^{(1)}\sum_{s=1}^{N_{z^{(1)}}} W_{ls}^{(1)} \phi\left({z'}^{(1)}_s(x_j)\right) \right] = \mathbb{E}\left[b_k^{(1)}\right] \sum_{s=1}^{N_{z^{(1)}}} \mathbb{E} \left[ W_{ls}^{(1)} \phi\left({z'}^{(1)}_s(x_j)\right) \right] = 0 \\
& \textstyle \mathbb{E}\left[ b_l^{(1)}\sum_{r=1}^{N_{z^{(1)}}} W_{kr}^{(1)} \phi\left({z'}^{(1)}_r(x_i)\right) \right] = \mathbb{E}\left[b_l^{(1)}\right] \sum_{r=1}^{N_{z^{(1)}}} \mathbb{E} \left[ W_{kr}^{(1)} \phi\left({z'}^{(1)}_r(x_i)\right) \right] = 0 \\
& \textstyle \mathbb{E}\left[ \left( \sum_{r=1}^{N_{z^{(1)}}} W_{kr}^{(1)} \phi\left({z'}^{(1)}_r(x_i)\right) \right) \left( \sum_{s=1}^{N_{z^{(1)}}} W_{ls}^{(1)} \phi\left({z'}^{(1)}_s(x_j)\right) \right) \right] \\
&= \sum_{r=1}^{N_{z^{(1)}}} \sum_{s=1}^{N_{z^{(1)}}} \mathbb{E} \left[ W_{kr}^{(1)} W_{ls}^{(1)} \phi\left({z'}^{(1)}_r(x_i)\right) 
\phi\left({z'}^{(1)}_s(x_j)\right) \right] \\
&= \sum_{r=1}^{N_{z^{(1)}}} \sum_{s=1}^{N_{z^{(1)}}} \mathbb{E} \left[ W_{kr}^{(1)} \right] \mathbb{E} \left[ W_{ls}^{(1)} \right] \mathbb{E} \left[ \phi\left({z'}^{(1)}_r(x_i)\right) 
\phi\left({z'}^{(1)}_s(x_j)\right) \right] \\
&= 0.
\end{align*}
Therefore, we have $\text{cov}\left( z_k^{(1)}(x_i), z_l^{(1)}(x_j) \right) = 0$, meaning that $z_{ik}^{(1)}$ and $z_{jl}^{(1)}$ are uncorrelated with each other \emph{a priori}, which completes the proof of Lemma \ref{lemma:2}. However, we do not claim the validity of this lemma \emph{a posteriori}.

\section{A More Detailed Proof for Theorem 1}
\renewcommand\thefigure{\thesection\arabic{figure}}
\setcounter{figure}{0}
\label{appx:B}

In this section, we provide a more detailed proof for Theorem 1. Again, suppose we have data $\boldsymbol{X} = \left[\boldsymbol{x}_i\right]_{i=1}^N$ with information from 2 different sources $\boldsymbol{x} = \left\{ \boldsymbol{x}^{(1)}, \boldsymbol{x}^{(2)} \right\}$ where $\boldsymbol{x}^{(1)} \in \mathbb{R}^{D_1}$ is high-dimensional and $\boldsymbol{x}^{(2)} \in \mathbb{R}^{D_2}$ is low-dimensional with some known relationship with the targets $\boldsymbol{Y} = [y_i]_{i=1}^N$. With ICK formulation, we have $\boldsymbol{z}^{(1)} = f_{\text{NN}} \left(\boldsymbol{x}^{(1)}\right) \in \mathbb{R}^p$ where $f_{\text{NN}}$ is a NN-implied function with parameters $\boldsymbol{\theta}^{(1)}$ and $\boldsymbol{z}^{(2)} = g\left(\boldsymbol{x}^{(2)}\right) \in \mathbb{R}^p$ where $g$ is a kernel-to-latent-space mapping specified by a chosen kernel function $K^{(2)}$ with parameters $\boldsymbol{\theta}^{(2)}$. In Section \ref{sec:3.3}, it is stated that the latent representation $\boldsymbol{z}^{(1)}$ from $f_{\text{NN}}$ will converge in distribution to a multi-output GP in the infinite width limit. In case when the NN has finite width, $\boldsymbol{z}^{(1)}$ will \emph{approximately} follows a GP with empirical NNGP kernel function $\hat{K}^{\text{NNGP}}$
\begin{equation}
\boldsymbol{z}^{(1)} \sim \mathcal{GP}_{\textit{approx}} \left(0, \hat{K}^{\text{NNGP}}\right).
\end{equation}
Let $\boldsymbol{Z}^{(1)} = \left[\boldsymbol{z}_i^{(1)}\right]_{i=1}^N$. Based on Lemma \ref{lemma:2}, $\boldsymbol{Z}^{(1)}$ can be approximately treated as a sample from a multivariate Gaussian as shown below
\begin{equation}
\boldsymbol{Z}^{(1)} \sim \mathcal{N} \left( \boldsymbol{0}, \hat{\boldsymbol{K}}^{\text{NNGP}} \right),
\end{equation}
where $\boldsymbol{Z}^{(1)} \in \mathbb{R}^{N \times p}$ and $\hat{\boldsymbol{K}}^{\text{NNGP}} \in \mathbb{R}^{Np \times Np}$ is the corresponding kernel matrix
\begin{equation}
\hat{\boldsymbol{K}}^{\text{NNGP}} = 
\begin{bmatrix}
\hat{\boldsymbol{K}}^{\text{NNGP}}_1 & \boldsymbol{0} & \cdots & \boldsymbol{0} \\
\boldsymbol{0} & \hat{\boldsymbol{K}}^{\text{NNGP}}_2 & \cdots & \boldsymbol{0} \\
\vdots & \vdots & \ddots & \vdots \\
\boldsymbol{0} & \boldsymbol{0} & \cdots & \hat{\boldsymbol{K}}^{\text{NNGP}}_p
\end{bmatrix}, 
\label{eqn:19}
\end{equation}
and $\hat{\boldsymbol{K}}^{\text{NNGP}}_k = \left[ \hat{K}^{\text{NNGP}}_k\left(x^{(1)}_i, x^{(1)}_j\right) \right]_{i,j=1,...,N} \in \mathbb{R}^{N \times N}$ for all $k = 1, 2, ..., p$. Also, since all entries in $\boldsymbol{z}^{(1)}$ are from the same NN architecture with parameters drawn from the same distribution, as we increase the NN width, all $\hat{K}^{\text{NNGP}}_1, ..., \hat{K}^{\text{NNGP}}_p$ will converge to a deterministic limit. Therefore, in the finite-width case, it is reasonable to say that $\hat{K}^{\text{NNGP}}_1 \approx \cdots \approx \hat{K}^{\text{NNGP}}_p \approx \hat{K}^{\text{NNGP}}$. Based on the marginalization property of GPs \citep{williams2006gaussian}, any $\boldsymbol{z}_i^{(1)} \in \mathbb{R}^p$ in $\boldsymbol{Z}^{(1)}$ should \emph{approximately} satisfy
\begin{equation}
\boldsymbol{z}_i^{(1)} \sim \mathcal{N} \left(\boldsymbol{0}, \hat{K}^{\text{NNGP}}\left(\boldsymbol{x}_i^{(1)}, \boldsymbol{x}_i^{(1)}\right)\boldsymbol{I}_p\right),
\end{equation}
where $\boldsymbol{I}_p$ is a $p \times p$ identity matrix. In other words, each entry in $\boldsymbol{z}_i^{(1)}$ has a univariate Gaussian distribution
\begin{equation}
z_{ik}^{(1)} \sim \mathcal{N} \left(0, \hat{K}^{\text{NNGP}}\left(\boldsymbol{x}_i^{(1)}, \boldsymbol{x}_i^{(1)}\right)\right), k = 1, 2, ..., p.
\end{equation}
Let $\alpha_{ik} = z_{ik}^{(2)} = g\left(\boldsymbol{x}_i^{(2)}\right)_k$, since the final output is $\hat{y} = f_{\text{ICK}} \left( \boldsymbol{x}^{(1)}, \boldsymbol{x}^{(2)} \right) = {\boldsymbol{z}^{(1)}}^T \boldsymbol{z}^{(2)}$, for the $i^{th}$ data point, $\hat{y}_i$ can be viewed as a weighted sum of $p$ independent Gaussian random variables under Lemma \ref{lemma:2}
\begin{align}
\hat{y}_i &\sim \sum_{k=1}^{p} \alpha_{ik} \mathcal{N} \left( 0, \hat{K}^{\text{NNGP}}\left(\boldsymbol{x}_i^{(1)}, \boldsymbol{x}_i^{(1)} \right) \right) \nonumber \\ 
&= \sum_{k=1}^p \mathcal{N} \left( 0, \alpha_{ik}^2 \hat{K}^{\text{NNGP}}\left(\boldsymbol{x}_i^{(1)}, \boldsymbol{x}_i^{(1)} \right) \right) \\ 
&= \mathcal{N} \left( 0, \hat{K}^{\text{NNGP}}\left(\boldsymbol{x}_i^{(1)}, \boldsymbol{x}_i^{(1)} \right) \sum_{k=1}^p \alpha_{ik}^2 \right).
\label{eqn:22}
\end{align}
Therefore, the joint distribution of the final predictions of the whole training set $\hat{\boldsymbol{Y}} = [\hat{y}_i]_{i=1}^N$ will be a multivariate Gaussian as given below:
\begin{equation}
\hat{\boldsymbol{Y}} \sim \mathcal{N} \left( \boldsymbol{0}, 
\sum_{k=1}^{p} \boldsymbol{\alpha}_k \boldsymbol{\alpha}_k^T
* \hat{\boldsymbol{K}}^{\text{NNGP}}
\right) = \mathcal{N} \left( \boldsymbol{0}, \boldsymbol{\alpha} \boldsymbol{\alpha}^T \odot \hat{\boldsymbol{K}}^{\text{NNGP}} \right),
\end{equation}
where $\boldsymbol{\alpha}_k = [\alpha_{ik}]_{i=1}^N \in \mathbb{R}^N$, $\boldsymbol{\alpha} = [\alpha_{ik}]_{\substack{i=1,...,N \\ k=1,...,p}} \in \mathbb{R}^{N \times p}$, $\hat{\boldsymbol{K}}^{\text{NNGP}} = \left[ \hat{K}^{\text{NNGP}} \left( \boldsymbol{x}_i^{(1)}, \boldsymbol{x}_j^{(1)} \right) \right]_{i,j = 1, ..., N}$, and "$\odot$" represents elementwise multiplication. Therefore, if $g$ includes the kernel-to-latent-space mapping $K_2 \left( \boldsymbol{x}_i^{(2)}, \boldsymbol{x}_j^{(2)} \right) \approx {\boldsymbol{z}_i^{(2)}}^T \boldsymbol{z}_j^{(2)} = g\left(\boldsymbol{x}_i^{(2)}\right)^T g\left(\boldsymbol{x}_j^{(2)}\right), i, j = 1, ..., N$ as shown in Equation \ref{eq:gapprox}, then $\boldsymbol{K}^{(2)} = \boldsymbol{\alpha} \boldsymbol{\alpha}^T$ and we derive that
\begin{equation}
\hat{\boldsymbol{Y}} \sim \mathcal{N} \left( \boldsymbol{0}, \hat{\boldsymbol{K}}^{\text{NNGP}} \odot \boldsymbol{K}_2 \right).
\end{equation}
Since this derivation reasonably holds for any finite set of $\boldsymbol{X}$ and $\boldsymbol{Y}$, we say the ICK framework is \emph{approximately} equivalent to a GP with zero mean and a multiplicative kernel between the NNGP kernel matrix $\hat{\boldsymbol{K}}^{\text{NNGP}}$ and the user-specified kernel matrix $\boldsymbol{K}_2$.

\section{Experimental Results of Random Fourier Features}
\renewcommand\thefigure{\thesection\arabic{figure}}
\setcounter{figure}{0}
\label{appx:C}

\subsection{Synthetic Data}
\label{appx:C1}
We use the same toy data set where each data point $\boldsymbol{x} = \left\{ x^{(1)}, x^{(2)}, x^{(3)} \right\}$ contains 3 sources of information as described in Section \ref{sec:5.1.2}. Also, we use the same types of kernels as those in ICK\emph{y}. The only difference here is that we use RFF instead of Nystr\"{o}m method to transform the kernel matrix into the latent space in ICK\emph{r} framework. 

\begin{figure}[t!]
\centering
\subfloat[\label{fig:C1a}]{
\includegraphics[width=0.32\linewidth]{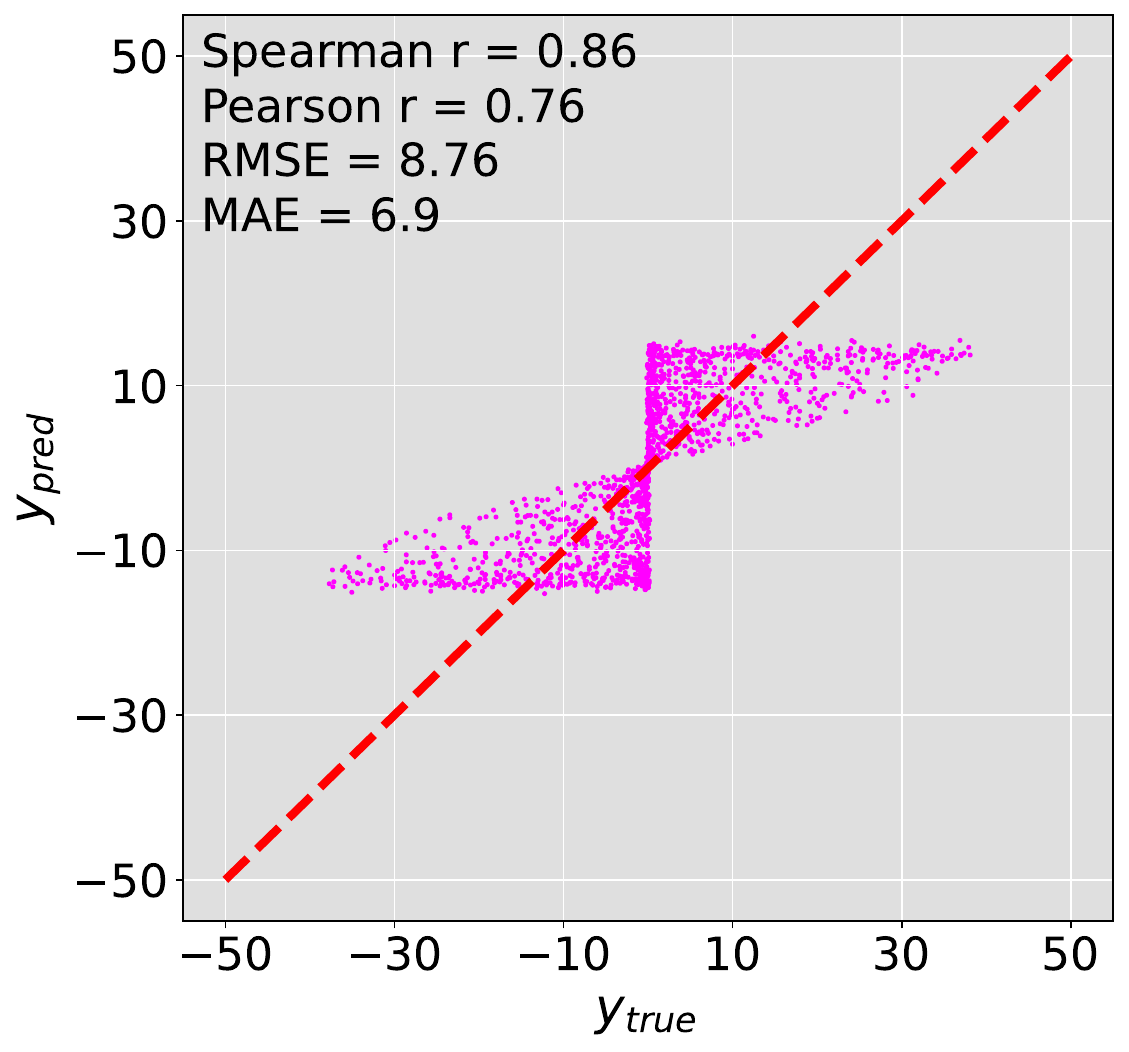}}
\subfloat[\label{fig:C1b}]{
\includegraphics[width=0.32\linewidth]{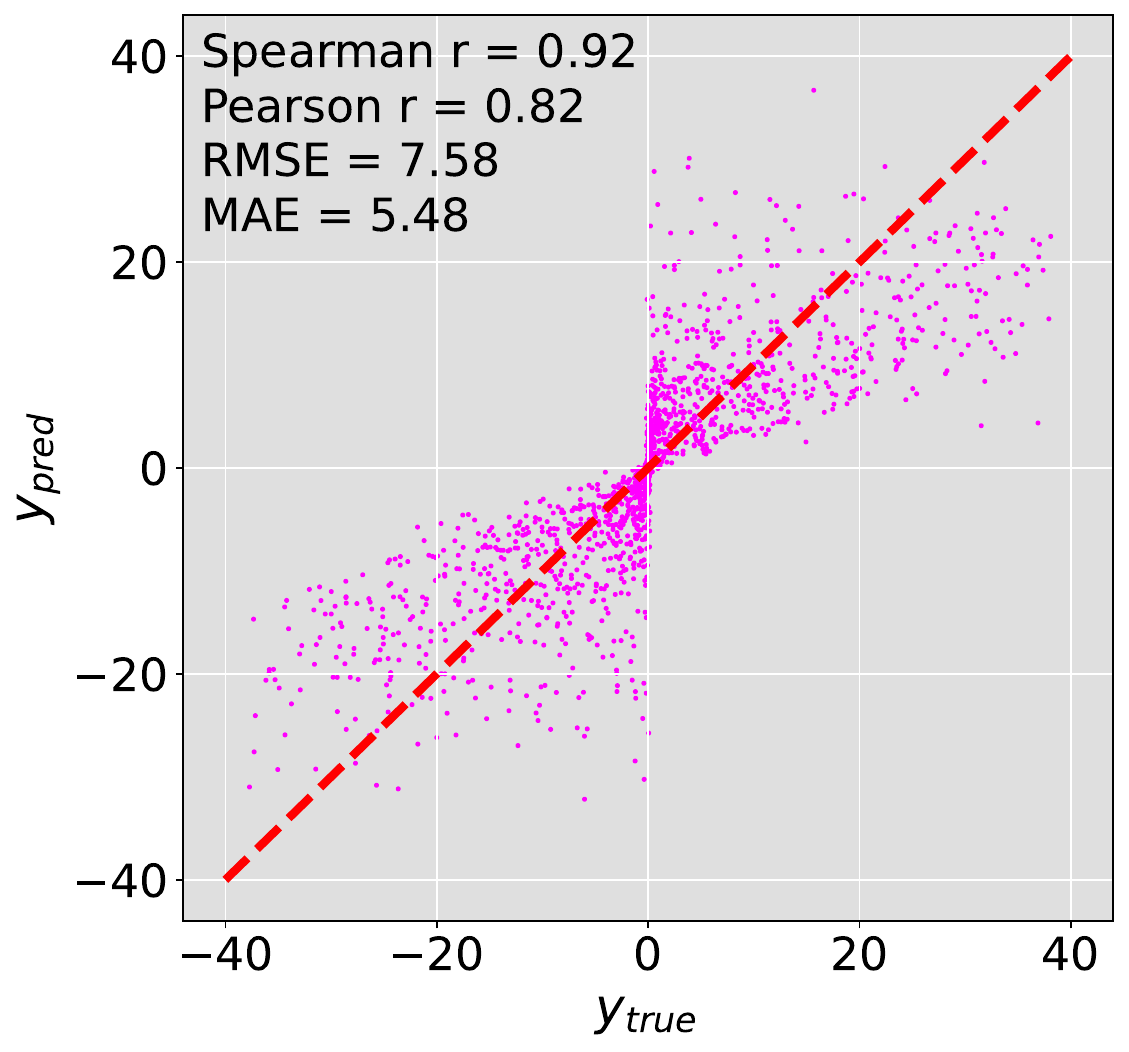}}
\subfloat[\label{fig:C1c}]{
\includegraphics[width=0.32\linewidth]{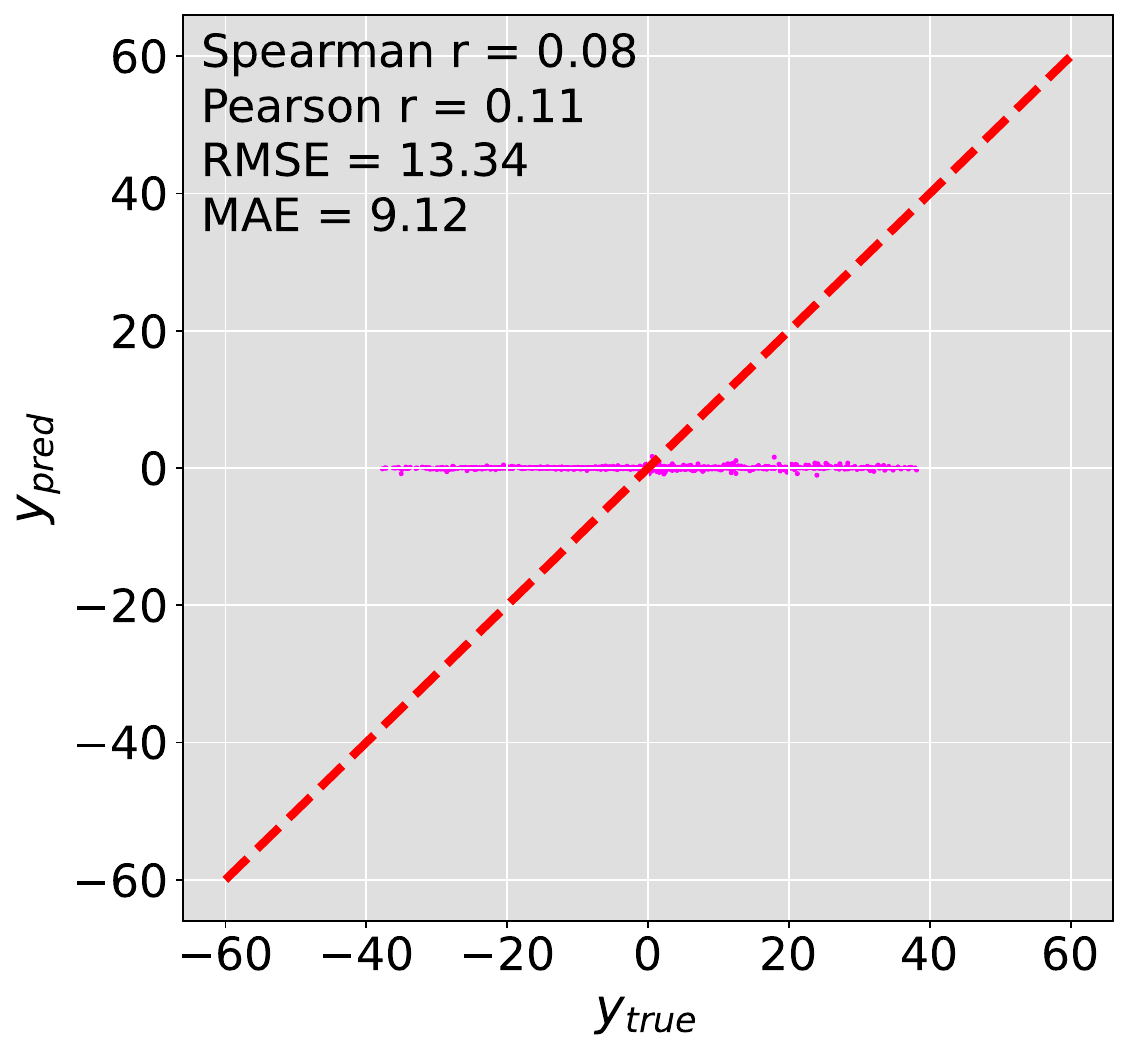}}
\caption{Scatter plots of the true values of $y$ against the predicted values of $y$ using our ICK\emph{r} framework with information from (a) one source $\hat{y} = f_{\text{NN}}\left(x^{(1)}\right)$, (b) 2 sources $\hat{y} = f_{\text{ICK\emph{r}}}\left(x^{(1)}, x^{(2)}\right)$, and (c) 3 sources $\hat{y} = f_{\text{ICK\emph{r}}}\left(x^{(1)}, x^{(2)}, x^{(3)}\right)$.}
\label{fig:C1}
\end{figure}

The results are displayed in Figure \ref{fig:C1}. It can be observed that when we add in only the side information $x^{(2)}$ along with the \emph{exponential sine squared} kernel, both the correlation and the predictive performance are improved (though not as good as the results from ICK\emph{y} as shown in Figure \ref{fig:9}). However, after we further include $x^{(3)}$ with the \emph{RBF} kernel, we realize that the parameters of ICK\emph{r} become very hard to optimize and it fails to make valid predictions and starts to guess randomly around zero.

\subsection{Remote Sensing Data}
\label{appx:C2}

\begin{figure}[t!]
\centering
\subfloat[\label{fig:C2a}]{
\includegraphics[width=0.48\linewidth]{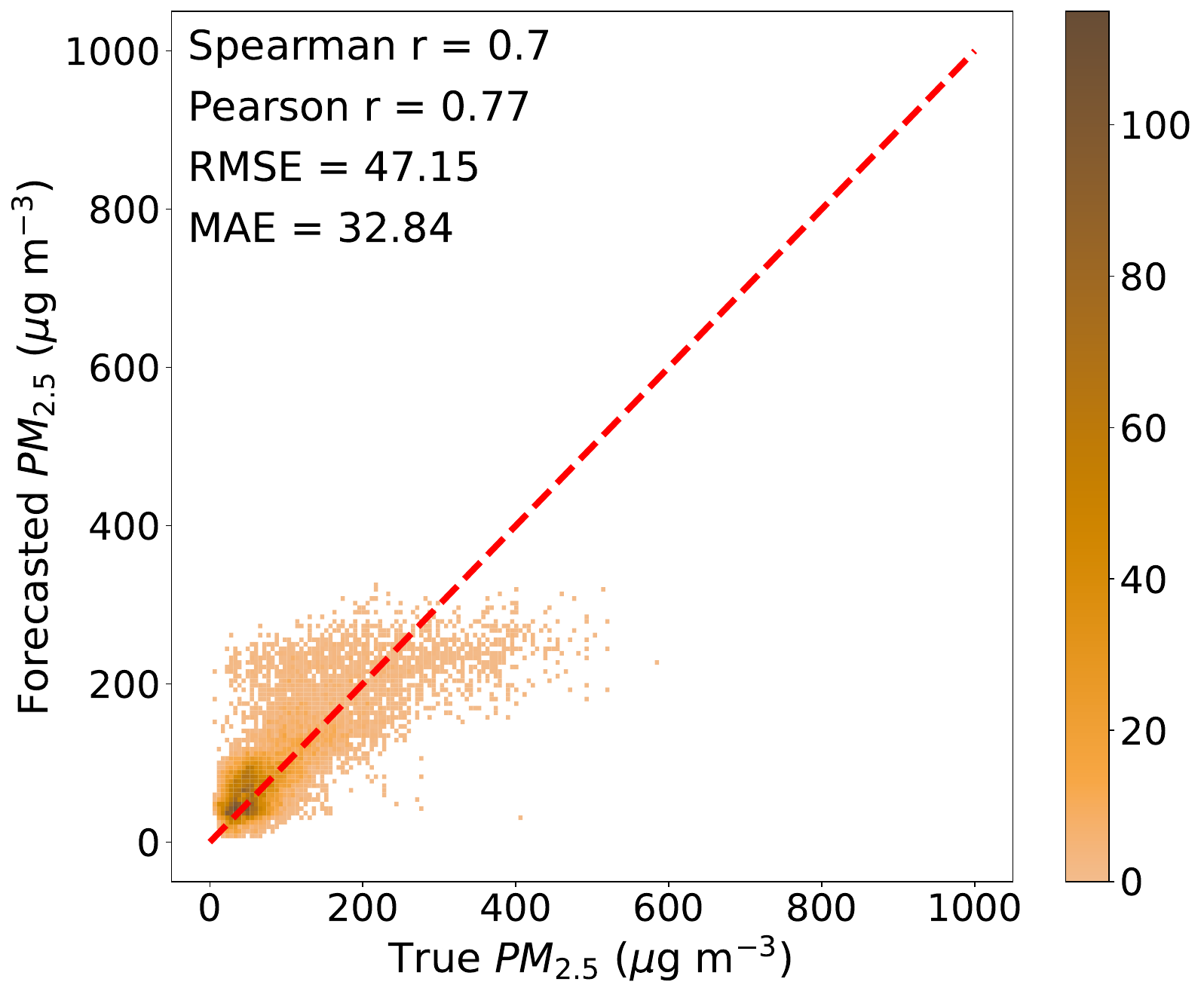}}
\hfill
\subfloat[\label{fig:C2b}]{
\includegraphics[width=0.48\linewidth]{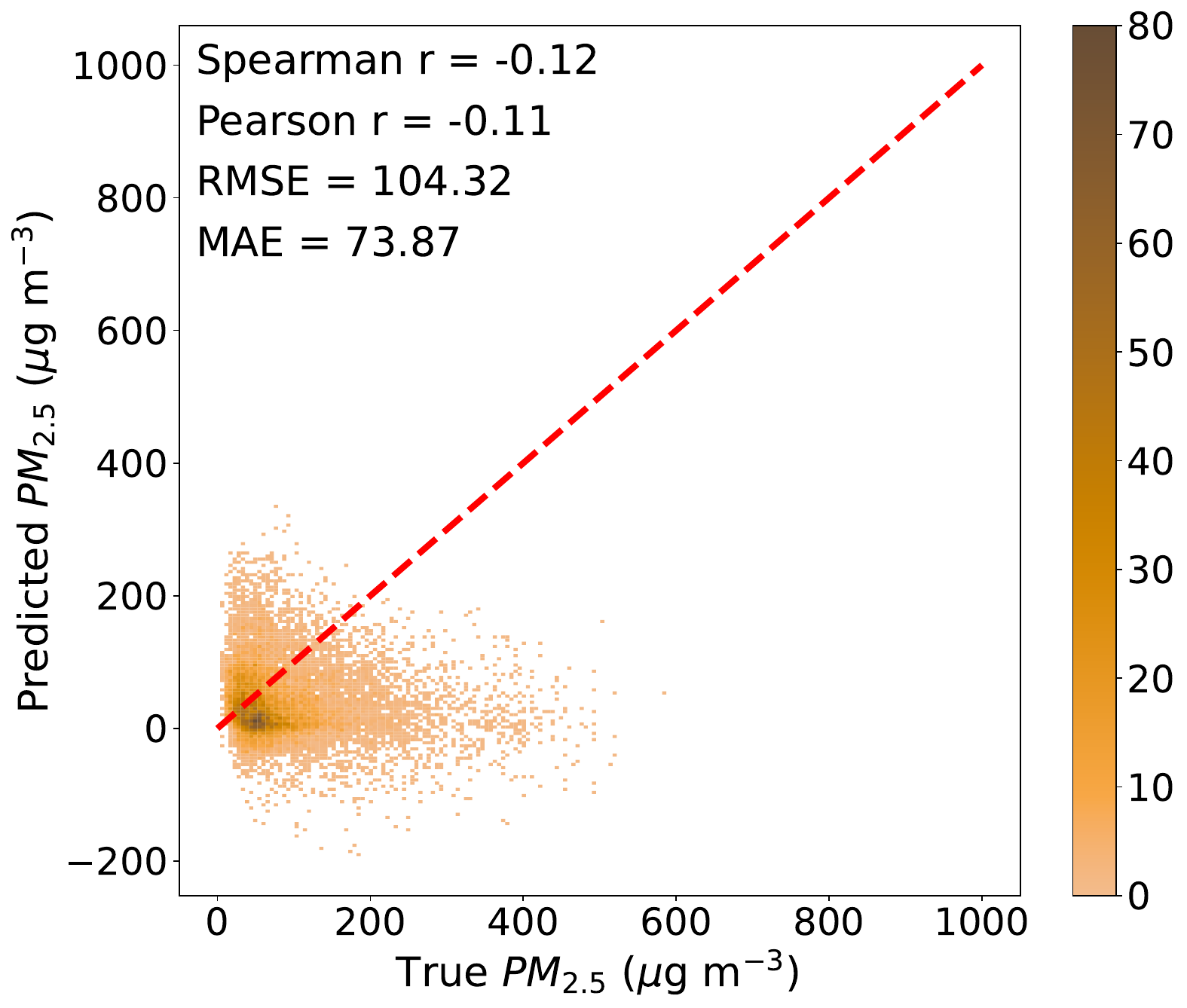}}
\caption{Density plots of the true PM$_{2.5}$ concentrations against the forecasted PM$_{2.5}$ concentrations for $t \geq 500$ using (a) ICK\emph{y} and (b) ICK\emph{r}.}
\label{fig:C2}
\end{figure}

We also try ICK\emph{r} on the \emph{forecasting} task using the remote sensing data (see Section \ref{sec:5.2}) and compare the results with those from ICK\emph{y}. Each data point $\boldsymbol{x} = \{x, t\}$ contains a satellite image $x$ as the high-dimensional information and its corresponding timestamp $t$ as the low-dimensional information. The satellite images are processed with a two-layer CNN and the timestamps are processed with an \emph{exponential-sine-squared} kernel with a period of $T = 365$ (days). As can be observed from Figure \ref{fig:C2}, ICK\emph{r} yields much higher error compared to ICK\emph{y}.

\section{Number of Inducing Points}
\renewcommand\thefigure{\thesection\arabic{figure}}
\setcounter{figure}{0}
\label{appx:D}

\begin{figure}[t!]
\centering
\subfloat[\label{fig:D1a}]{
\includegraphics[width=0.48\linewidth]{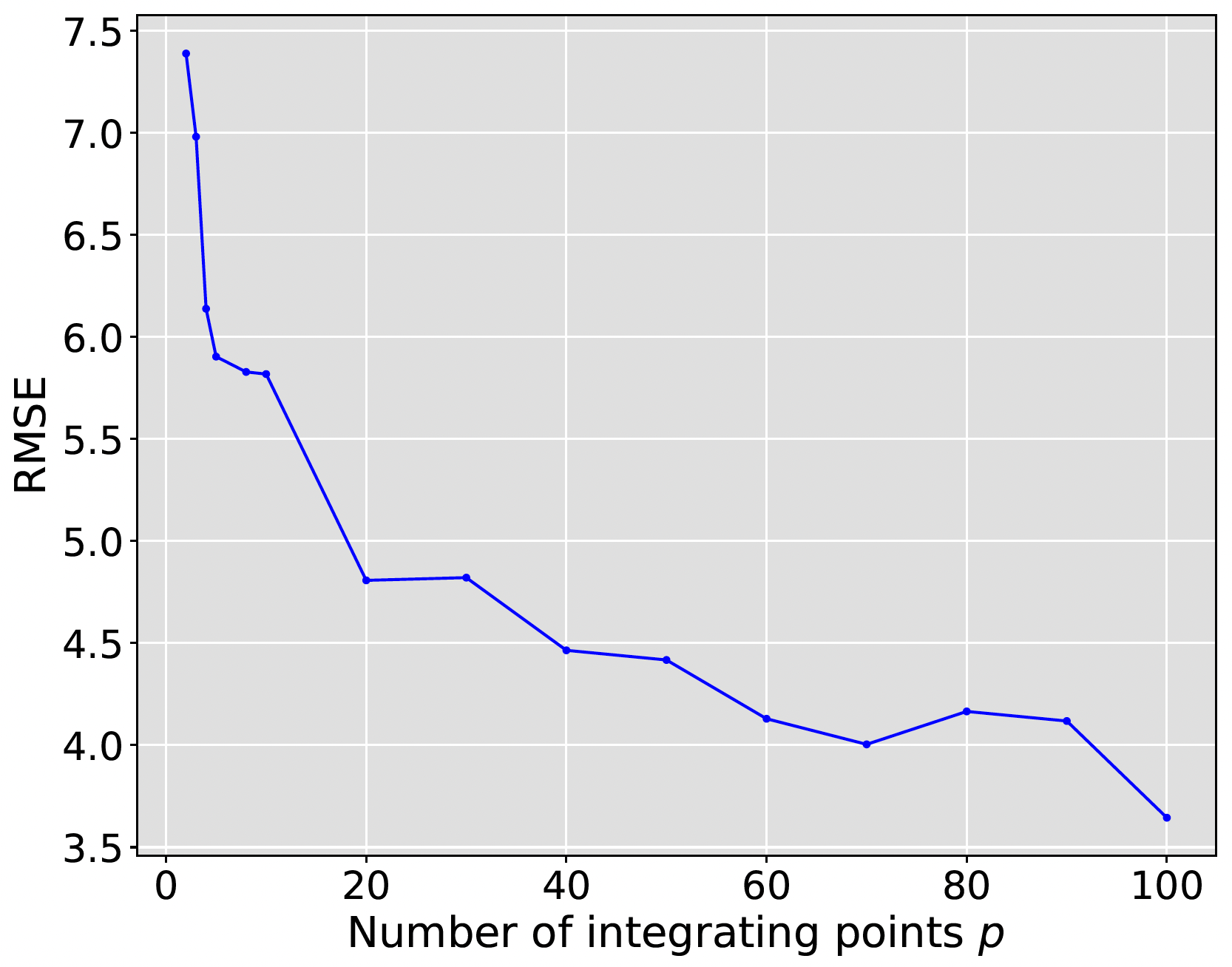}}
\hfill
\subfloat[\label{fig:D1b}]{
\includegraphics[width=0.48\linewidth]{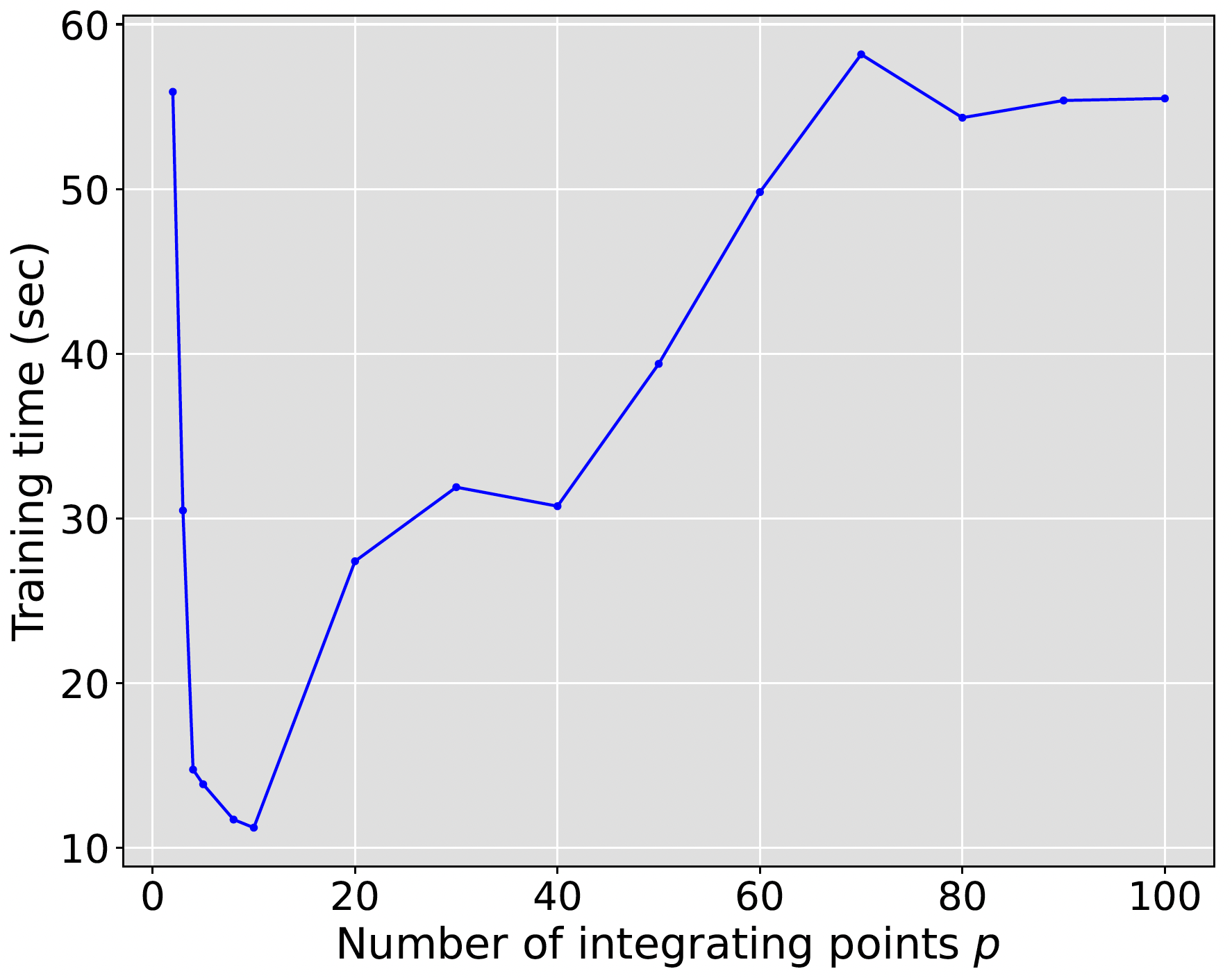}}
\caption{Plots of (a) prediction error and (b) training time of $\hat{y} = f_{\text{ICK\emph{y}}}\left(x^{(1)}, x^{(2)}, x^{(3)}\right)$ against the number of inducing points $p$}
\label{fig:D1}
\end{figure}

As discussed in Section \ref{sec:4.2.1}, as we increase the number of inducing points $p$, we expect the approximation error between the true kernel matrix $\boldsymbol{K}$ and the approximated kernel matrix $\hat{\boldsymbol{K}}$ to decrease.  Here, we empirically show how the value of $p$ impacts our predictions. In Figure \ref{fig:D1a}, we plot the prediction error of $\hat{y} = f_{\text{ICK\emph{y}}}\left(x^{(1)}, x^{(2)}, x^{(3)}\right)$ against the number of inducing points using the synthetic data generated in Appendix \ref{appx:F}. As can be observed, the prediction error drops sharply as we raise $p$ from a small value (e.g. $p = 2$). When $p$ is relatively large, increasing $p$ yields smaller improvement on the predictions. Additionally, in Figure \ref{fig:D1b}, we plot the total training time against $p$. The total training time is dependent on how long a single iteration takes and the total number of epochs required.  We note that once $p>80$ the training time is relatively flat, which is due to the fact that the total computation in the Cholesky is less than the computation in the neural network. Interestingly, it appears that when $p$ is very small, ICK\emph{y} takes longer to converge due to the need for many more epochs. As we increase $p$, the training time goes down and then goes up again due to the computational complexity, i.e. $\mathcal{O}(p^3)$, of the Cholesky decomposition. Based on these observations, we are not concerned about the computational complexity for reasonable values of $p$.

\section{Applying Sample-then-optimize Procedure to ICK}
\renewcommand\thefigure{\thesection\arabic{figure}}
\setcounter{figure}{0}
\label{appx:E}

As elaborated in Section \ref{sec:4.3.1}, a deep ensemble with proper initialization scheme will have a GP posterior interpretation in the infinite width limit when trained by a squared-error loss \citep{he2020bayesian}, which is an example of the "sample-then-optimize" procedure of \citet{matthews2017sample} Algorithm \ref{alg:3} is an alternative to replace $f_{\text{NN}}$ with such a deep ensemble $F = \left\{f_{n_e}\right\}_{n_e = 1}^{N_e}$ where the dimension of the final readout layer of each $f_{n_e}$ is $p$. In this case, each baselearner $f_{n_e}$ can be viewed as an i.i.d. sample from a multi-output GP in the infinite width limit. In the finite-width case, this relationship becomes approximate:
\begin{equation}
f_{n_e}  \xrightarrow[]{d} \mathcal{GP}_{\textit{approx}} \left(0, K_F\right),
\end{equation}
As stated by \citet{lee2019wide}, if all the parameters in $f_{n_e}$ are randomly drawn from a Gaussian distribution and are all fixed except the last layer, then after training $F$ on a squared-error loss, we will have $K_F \rightarrow K^{\text{NNGP}}$ where $K^{\text{NNGP}}$ is the NNGP kernel. \citet{he2020bayesian} also proposed to add a random and untrainable function $\delta(\cdot)$ to the output of $f_{n_e}$ so we have $K_F \rightarrow K^{\text{NTK}}$ where $K^{\text{NTK}}$ is the limiting NTK.

With ICK formulation, by following the proof given in Appendix \ref{appx:B}, we can derive that, for each baselearner in the ICK ensemble, the joint distribution of the final predictions of the whole training set $\hat{\boldsymbol{Y}}_{n_e}$ will again be a multivariate Gaussian as shown below if Lemma \ref{lemma:2} holds \emph{a posteriori}:
\begin{equation}
\hat{\boldsymbol{Y}}_{n_e} \sim \mathcal{N} \left( \boldsymbol{0}, \boldsymbol{K}_F \odot \boldsymbol{K}^{(2)} \right),
\label{eqn:32}
\end{equation}
where $\left(\boldsymbol{K}_F\right)_{ij} = K_F\left(\boldsymbol{x}_i^{(1)}, \boldsymbol{x}_j^{(1)}\right)$. Since Equation \ref{eqn:32} holds for any finite input data set, we can conclude that each baselearner $f_s$ in the ICK ensemble $F_{\text{ICK}y} = \{f_s\}_{s=1}^{N_e}$ can be \emph{approximately} viewed as an i.i.d. sample from a single-output GP
\begin{equation}
f_s \sim \mathcal{GP}_{\textit{approx}} \left( 0, K_F K^{(2)} \right) = \mathcal{GP}_{\textit{approx}} \left( 0, K^{\text{comp}} \right).
\end{equation}
Building an ensemble $F_{\text{ICK}y} = \{f_s\}_{s=1}^{N_e}$ is thus equivalent to performing a Monte Carlo approximation to a GP predictive posterior distribution whose mean and covariance matrix for a test data set $\boldsymbol{X}^* = \left[\boldsymbol{x}_i^*\right]_{i=1}^{N^*}$ are
\begin{align}
f_s (\boldsymbol{X}^*|\boldsymbol{X}) &\sim \mathcal{N} \left( \boldsymbol{\mu}^*, \boldsymbol{\Sigma}^* \right), \\
\boldsymbol{\mu}^* &= \boldsymbol{K}^{\text{comp}}_{X^*X} \left(\boldsymbol{K}^{\text{comp}}_{XX}\right)^{-1} \boldsymbol{Y}, \\
\boldsymbol{\Sigma}^* &= \boldsymbol{K}^{\text{comp}}_{X^* X^*} - \boldsymbol{K}^{\text{comp}}_{X^*X} \left(\boldsymbol{K}^{\text{comp}}_{XX}\right)^{-1} \boldsymbol{K}^{\text{comp}}_{XX^*},
\end{align}
where $\left(\boldsymbol{K}^{\text{comp}}_{X^*X}\right)_{ij} = K^{\text{comp}}\left(\boldsymbol{x}_i^*, \boldsymbol{x}_j\right)$, $\left(\boldsymbol{K}^{\text{comp}}_{XX}\right)_{ij} = K^{\text{comp}}\left(\boldsymbol{x}_i, \boldsymbol{x}_j\right)$, $\left(\boldsymbol{K}^{\text{comp}}_{X^* X^*}\right)_{ij} = K^{\text{comp}}\left(\boldsymbol{x}_i^*, \boldsymbol{x}_j^*\right)$, and $\boldsymbol{K}^{\text{comp}}_{XX^*} = \left(\boldsymbol{K}^{\text{comp}}_{X^*X}\right)^T$. In other words, we can approximate the GP predictive posterior using the predictive mean and variance generated by Algorithm \ref{alg:3} as follows:
\begin{align}
\hat{\boldsymbol{\mu}}^* &= \frac{1}{N_e} \sum_{n=1}^{N_e} f_s (\boldsymbol{X}^*|\boldsymbol{X}) \approx \mathbb{E}\left[ f_s (\boldsymbol{X}^*|\boldsymbol{X}) \right] = \boldsymbol{\mu}^*, \\
{\hat{\boldsymbol{\sigma}}}^{*2} &= \frac{1}{N_e} \sum_{n=1}^{N_e} \left[ f_s (\boldsymbol{X}^*|\boldsymbol{X}) - \hat{\boldsymbol{\mu}}^* \right]^2 \approx \mathbb{V} \left[ f_s (\boldsymbol{X}^*|\boldsymbol{X}) \right] \nonumber \\ 
&= \text{diag} \left( \boldsymbol{\Sigma}^* \right). 
\end{align}

\section{Experimental Details}
\label{appx:F}
\renewcommand\thefigure{\thesection\arabic{figure}}
\setcounter{figure}{0}

\begin{table}
\setlength{\tabcolsep}{0.3em}
\centering
\caption{Model architecture and training details for remote sensing data experiment in Section \ref{sec:5.2}}
\begin{tabular}{c|c|c|c} 
\hline
~            & \begin{tabular}[c]{@{}c@{}}Backbone architecture \\details\end{tabular}                                                                    & \begin{tabular}[c]{@{}c@{}}Output FC layers \\dimension\end{tabular} & Optimizer                                                    \\ 
\hline
CNN-RF       & \begin{tabular}[c]{@{}c@{}}\# Conv blocks = 2, \# Channels = 16, \\Kernel size = 3, Stride = 1\end{tabular}                                            & 1000 + $d_{\text{RT}}$, 512, 512, 1                                                    & \begin{tabular}[c]{@{}c@{}}Adam\\$\beta_1 = 0.9$ \\ $\beta_2 = 0.999$\end{tabular}                                                         \\ 
\hline
ViT-RF       & \begin{tabular}[c]{@{}c@{}}\# Transformer blocks = 2, \\\# Attention heads = 8, \\Dropout ratio = 0.1\end{tabular}                         & 1000 + $d_{\text{RT}}$, 512, 512, 1                                                    & \begin{tabular}[c]{@{}c@{}}Adam\\$\beta_1 = 0.9$ \\ $\beta_2 = 0.999$\end{tabular}                                                         \\ 
\hline
DeepViT-RF   & \begin{tabular}[c]{@{}c@{}}\# Transformer blocks = 2, \\\# Attention heads = 8, \\Dropout ratio = 0.1\end{tabular}                         & 1000 + $d_{\text{RT}}$, 512, 512, 1                                                    & \begin{tabular}[c]{@{}c@{}}Adam\\$\beta_1 = 0.9$ \\ $\beta_2 = 0.999$\end{tabular}                                                         \\ 
\hline
MAE-ViT-RF   & \begin{tabular}[c]{@{}c@{}}\# Transformer blocks = 2, \\\# Attention heads = 8, \\Dropout ratio = 0.1, \\Masking ratio = 0.75\end{tabular} & 1000 + $d_{\text{RT}}$, 512, 512, 1                                                    & \begin{tabular}[c]{@{}c@{}}Adam\\$\beta_1 = 0.9$ \\ $\beta_2 = 0.999$\end{tabular}                                                         \\ 
\hline\hline
CNN-ICKy     & \begin{tabular}[c]{@{}c@{}}\# Conv blocks = 2, \# Channels = 16, \\Kernel size = 3, Stride = 1\end{tabular}                                          & 1000, 512, $p$                                                         & \begin{tabular}[c]{@{}c@{}}SGD\\momentum = 0.9\end{tabular}  \\ 
\hline
ViT-ICKy     & \begin{tabular}[c]{@{}c@{}}\# Transformer blocks = 2, \\\# Attention heads = 8, \\Dropout ratio = 0.1\end{tabular}                         & 1000, 512, $p$                                                         & \begin{tabular}[c]{@{}c@{}}SGD\\momentum = 0.9\end{tabular}  \\
\hline
DeepViT-ICKy & \begin{tabular}[c]{@{}c@{}}\# Transformer blocks = 2, \\\# Attention heads = 8, \\Dropout ratio = 0.1\end{tabular}                         & 1000, 512, $p$                                                         & \begin{tabular}[c]{@{}c@{}}SGD\\momentum = 0.9\end{tabular}  \\
\hline
\end{tabular}
\label{tab:4}
\end{table}

\subsection{Synthetic Data}
\label{appx:F1}

We use the GPytorch package to generate the synthetic data. The data set is first randomly shuffled and then divided into train and test set with a 50:50 ratio.  Before feeding $x^{(1)}$ into MLP, we first map $x^{(1)}$ into higher dimension using an unsupervised algorithm called Totally Random Trees
Embedding. All the MLP structures in this experiment (including those in MLP-RF and ICK\emph{y}) contain one single fully connected (FC) layer of width 1000, which serves as a simple benchmark since a one-hidden-layer MLP can only capture linear relationship between the input and output. For model training, we optimize a Mean Squared Error (MSE) objective using Adam optimizer with a weight decay of 0.1.

\subsection{Remote Sensing Data}
\label{appx:F2}

We collect remote sensing data from 51 air quality monitoring (AQM) stations located in the National Capital Territory (NCT) of Delhi and its satellite cities over the period from January 1, 2018 to June 30, 2020 (see Appendix \ref{appx:G} for notes on data availability). The timestamps are converted into numerical values $t$ (where the day 2018-01-01 corresponds to $t = 0$) before feeding them into the models. We split the train, validation, and test data set based on $t$. Specifically, we use all the data points with $t < 365$ for training, $365 \leq t < 500$ for validation, and $t \geq 500$ for testing. 

The model architecture and training details are listed in Table \ref{tab:4}. Here $p$ denotes the length of latent representations $\boldsymbol{z}$ as discussed in Section \ref{sec:4} and $d_{\text{RT}}$ denotes the transformed dimension of timestamp $t$ using the Random Trees Embedding as mentioned in Section \ref{appx:F1}. Note that we use stochastic gradient descent (SGD) optimizer with a momentum of 0.9 for ICK\emph{y} as we realize that SGD helps ICK\emph{y} find a local minimum on the objective more efficiently. We use MSE objective for ICK\emph{y} and all benchmark models in this experiment.

\subsection{Other Regression Datasets}
\label{appx:F3}

For the worker productivity, we separate out the temporal information (i.e. date and time) and use it as the low-dimensional information. The rest of the features are then concatenated together to serve as the high-dimensional information. The MLPs (including the MLP part in ICK\emph{y}) in this experiment share the same structure as the one used in \citet{al2019deep}, which consist of 3 hidden layers of width 128, 32, and 32, respectively. For plain MLP, cyclic MLP, and ICK\emph{y}, we use the mean absolute error (MAE) objective to put less weight on the outliers and thus enhance the model performance. All these objectives are optimized by an Adam optimizer with $\beta_1 = 0.9$ and $\beta_2 = 0.999$.

The power consumption data is preprocessed in a similar way to the worker productivity data, where we separate out the data and time features, transform them into a one-dimensional time index, and use it as the low-dimensional information. The rest of the features are concatenated together to serve as the high-dimensional information. The performance of the GP benchmarks are directly obtained from \citet{wang2019exact}.

\section{Accessibility and Restrictions of the Data}
\label{appx:G}
\renewcommand\thefigure{\thesection\arabic{figure}}
\setcounter{figure}{0}

All experiments are conducted on a computer cluster equipped with a GeForce RTX 2080 Ti GPU. The synthetic data in Section \ref{sec:5.1} are generated using the GPyTorch package. The remote sensing data in Section \ref{sec:5.2} is downloaded using PlanetScope API whose content is protected by copyright and/or other intellectual property laws. To access the data on PlanetScope, the purchase of an end-user license is required.  When this manuscript is accepted, we will provide the codes we used to acquire the data. The UCI machine learning repository data we use in Section \ref{sec:5.3} has an open access license, meaning that the data is freely available online. 

\section{Time Series Visualization for Remote Sensing Experiment}
\label{appx:H}
\renewcommand\thefigure{\thesection\arabic{figure}}
\setcounter{figure}{0}

\begin{figure}[t!]
\centering
\subfloat[\label{fig:H1a}]{
\includegraphics[width=0.46\linewidth]{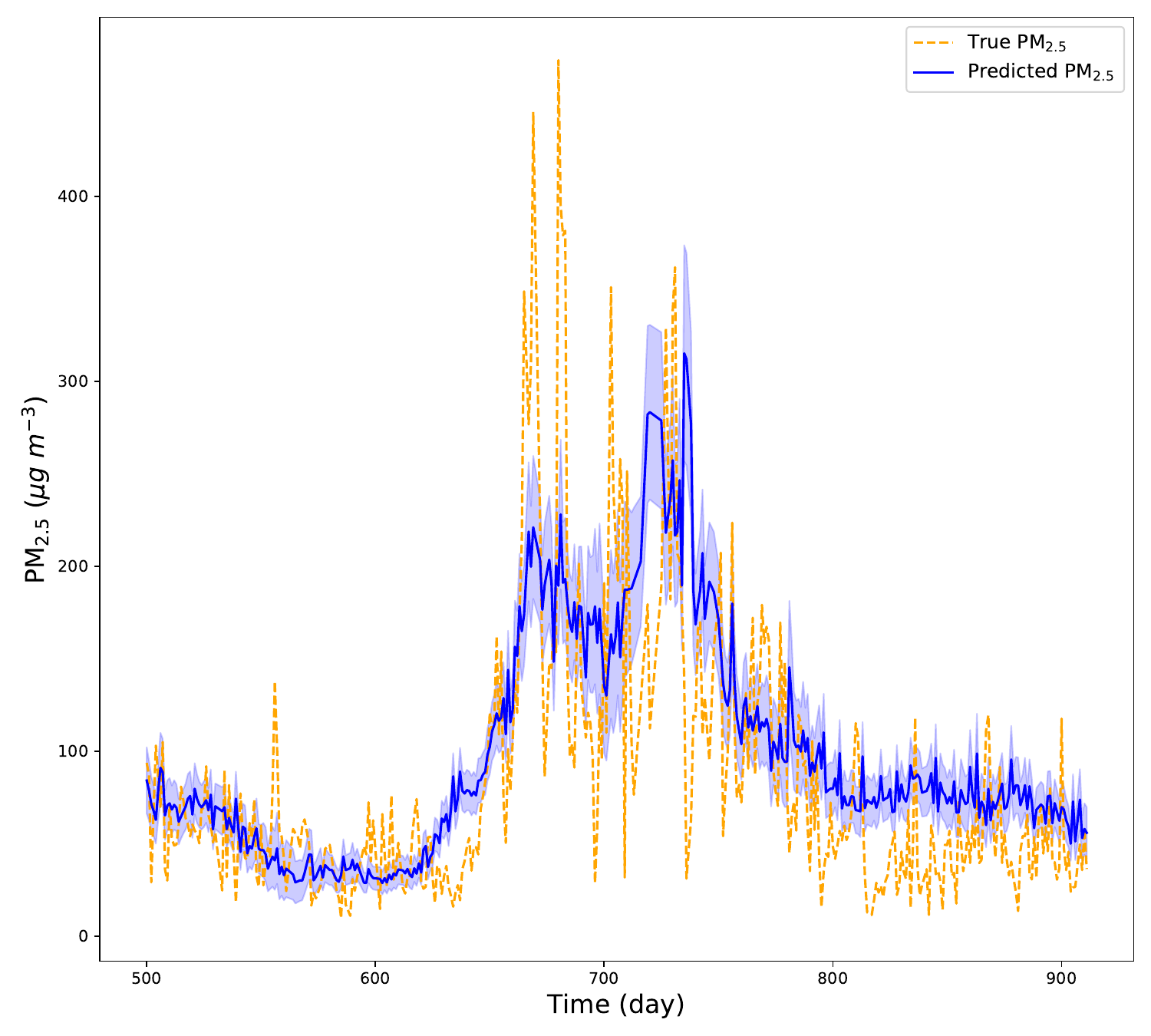}}
\hfill
\subfloat[\label{fig:H1b}]{
\includegraphics[width=0.46\linewidth]{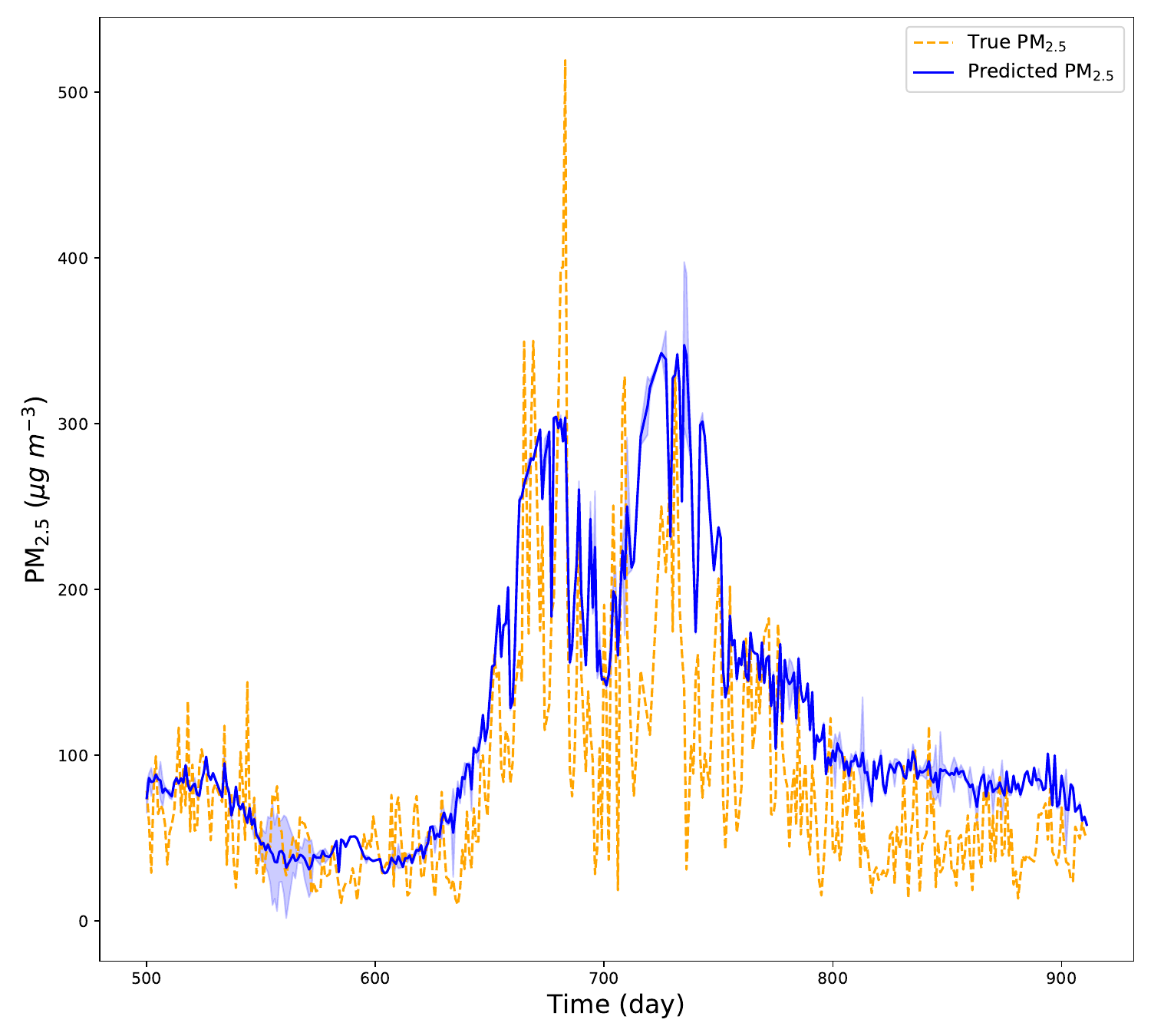}}
\caption{Time series visualization of predictive mean and uncertainty of PM$_{2.5}$ in Section \ref{sec:5.2} (remote sensing experiment) for (a) CNN-ICK\emph{y} (1.92M parameters) and (b) DeepViT-ICK\emph{y} (21.80M parameters).}
\label{fig:H1}
\vspace{-3mm}
\end{figure}

\textcolor{black}{To explore the problem of posterior uncertainty calibration for ViT variants of ICK\emph{y} (i.e. extremely large MSLL as shown in Table \ref{tab:1} in Section \ref{sec:5.2}), we visualize the results as time series by first grouping the predictions by the timestamp $t$, taking the \emph{minimum} of the predictive variance, and plotting them along with the corresponding true values and the predictive mean as shown in Figure \ref{fig:H1}. The shaded region represents the confidence interval $[\mu - 2\sigma, \mu + 2\sigma]$ where $\mu$ and $\sigma$ are the predictive mean and standard deviation of PM$_{2.5}$, respectively. We realize that CNN-ICK\emph{y} ensemble tends to yield much higher variance than DeepViT-ICK\emph{y} ensemble when the predictive mean deviates from the true values. A plausible explanation is that the DeepViT structure contains much more parameters than the CNN structure (21.80M vs 1.92M) used in our remote sensing experiment, which makes DeepViT-ICK\emph{y} overparameterized (The DeepViT architecture is set to be consistent with the CNN architecture as shown in Table \ref{tab:4}). To alleviate this problem, we try reducing the number of transformer blocks in ViT and DeepViT and we do observe a significant drop in MSLL.}

\end{document}